\documentclass{article}

\usepackage{microtype}
\usepackage{graphicx}
\usepackage{booktabs} 

\usepackage{hyperref}

\usepackage[accepted]{sysml2019}

\usepackage[utf8]{inputenc} 
\usepackage[T1]{fontenc}    
\usepackage{hyperref}       
\usepackage{url}            
\usepackage{booktabs}       
\usepackage{amsfonts}       
\usepackage{nicefrac}       
\usepackage{microtype}      
\usepackage{mathtools}
\DeclarePairedDelimiter{\ceil}{\lceil}{\rceil}
\DeclarePairedDelimiter\floor{\lfloor}{\rfloor}
\usepackage{fixltx2e}
\newcommand{\ssep}{\,;\,}
\newcommand{\param}{\vec{\theta}}
\newcommand{\fgemm}{\diamond}
\usepackage{algorithm}
\usepackage{algpseudocode}
\renewcommand{\algorithmicrequire}{\textbf{Input:}}
\renewcommand{\algorithmicensure}{\textbf{Output:}}
\algrenewcommand\alglinenumber[1]{\tiny #1:}

\algdef{SE}[VARIABLES]{Variables}{EndVariables}
   {\algorithmicvariables}
   {\algorithmicend\ \algorithmicvariables}
\algnewcommand{\algorithmicvariables}{\textbf{global variables}}
\algblockdefx{FORALLP}{ENDFAP}[1]%
  {\textbf{for all }#1 \textbf{do in parallel}}%
  {\textbf{end for}}
\usepackage{subcaption}
\usepackage{footnote}
\makesavenoteenv{tabular}
\makesavenoteenv{table}
\algrenewcommand\Return{\State \algorithmicreturn{} }
\usepackage[super]{nth}

\usepackage{enumitem}
\usepackage{threeparttable}
\usepackage{booktabs, caption, makecell}
\usepackage{titlesec}
\usepackage{bm}
\usepackage{amssymb}
\usepackage[normalem]{ulem}
\usepackage{multirow}

\usepackage[firstpage]{draftwatermark}
\SetWatermarkText{
 \hspace*{2.3in}
 \raisebox{10in}{
  \includegraphics[height=0.9in]{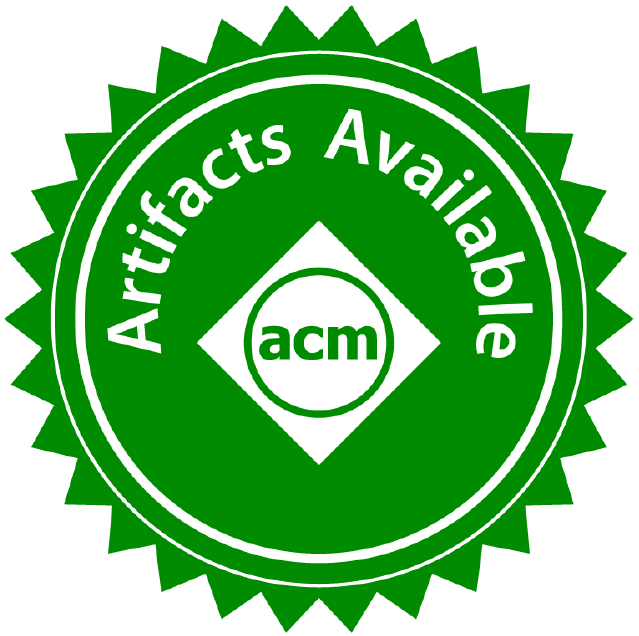}
  \includegraphics[height=0.9in]{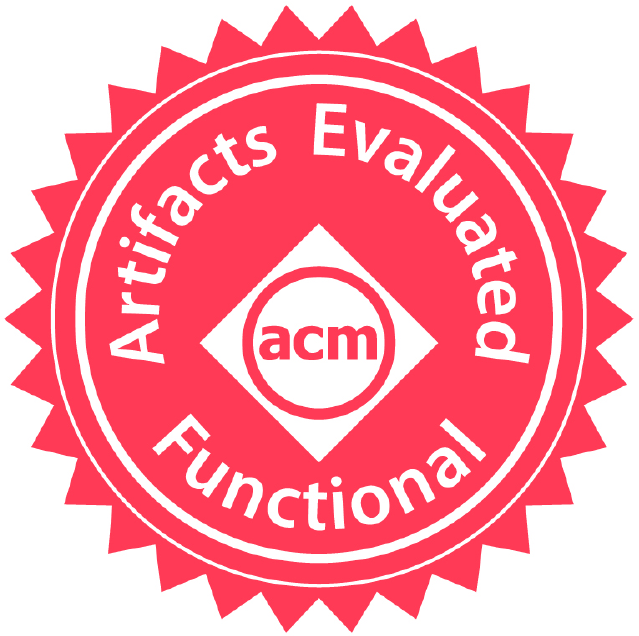}
  \includegraphics[height=0.9in]{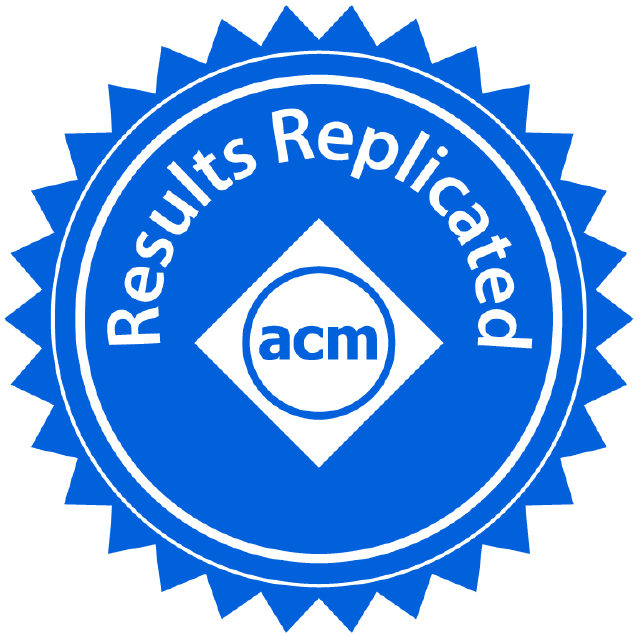}
 }
}
\SetWatermarkAngle{0}

\sysmltitlerunning{BPPSA: Scaling Back-propagation by Parallel Scan Algorithm}

\begin{document}

\twocolumn[
\sysmltitle{BPPSA: Scaling Back-propagation by Parallel Scan Algorithm}



\sysmlsetsymbol{equal}{*}

\begin{sysmlauthorlist}
\sysmlauthor{Shang Wang}{uoft}
\sysmlauthor{Yifan Bai}{berkely}
\sysmlauthor{Gennady Pekhimenko}{uoft}
\end{sysmlauthorlist}

\sysmlaffiliation{uoft}{Department of Computer Science, University of Toronto}
\sysmlaffiliation{berkely}{Department of Electrical Engineering and Computer Sciences, University of California, Berkeley; work done at the University of Toronto}

\sysmlcorrespondingauthor{Shang Wang}{wangsh46@cs.toronto.edu}

\sysmlkeywords{Machine Learning, SysML}

\vskip 0.3in

\vspace{-0.3cm}
\begin{abstract}
In an era when the performance of a single compute device plateaus, software must be designed to scale on massively parallel systems for better runtime performance. However, in the context of training deep learning models, the popular back-propagation (BP) algorithm imposes a \emph{strong sequential dependency} in the process of gradient computation. Under model parallelism, BP takes $\Theta (n)$ steps to complete which hinders its scalability on parallel systems ($n$ represents the number of compute devices into which a model is partitioned).

In this work, in order to improve the scalability of BP, we reformulate BP into a \emph{scan} operation which is a primitive that performs an in-order aggregation on a sequence of values and returns the partial result at each step. We can then scale such reformulation of BP on parallel systems by our modified version of the Blelloch scan algorithm which theoretically takes $\Theta (\log n)$ steps. We evaluate our approach on a vanilla Recurrent Neural Network (RNN) training with synthetic datasets and a RNN with Gated Recurrent Units (GRU) training with the IRMAS dataset, and demonstrate up to $2.75\times$ speedup on the overall training time and $108\times$ speedup on the backward pass. We also demonstrate that the retraining of pruned networks can be a practical use case of our method.
\end{abstract} \vspace{-0.2cm}
]



\printAffiliationsAndNotice{}  

\section{Introduction}

The training of deep learning models demands more and more compute resources \cite{openai_exp_high_ML_compute} as the models become more powerful and complex with an increasing number of layers in recent years \cite{alexnet, googlenet, vgg, resnet, densenet}. For example, ResNet can have more than a thousand layers \cite{identity_mapping}, and ResNet-152 takes days to train on eight state-of-the-art GPUs \cite{dawn_bench}. Now that the performance of a single compute device plateaus \cite{dark_silicon, mcm_gpu}, training has to be designed to scale on massively parallel systems.

Data parallelism \cite{data_parallelism} is the most popular way to scale training by partitioning the training data among multiple devices, where each device contains a full replica of the model. As the number of devices increases, data parallelism faces the trade-off between synchronization cost in synchronous parameter updates and staleness in asynchronous parameter updates \cite{demystifying_parallel_and_distributed_deep_learning}. Furthermore, recent studies demonstrate the scaling limit of data parallelism even when assuming perfect implementations and zero synchronization cost \cite{data_parallelism}. Lastly, data parallelism cannot be applied when a model does not fit into one device due to memory constraints (e.g., caused by deep network architecture, large batch size, or high input resolution \cite{vdnn, tbd}).

Model parallelism \cite{alex_model_parallelism, gpipe, mesh_tf, pipedream} is another approach to distributed training which partitions a model and distributes its parts among devices. It covers a wide spectrum of training deep learning models where data parallelism does not suffice. Na\"ive training under model parallelism does not scale well with the number of devices due to under-utilization of the hardware resources, since \emph{at most one} device can be utilized at any given point in time \cite{pipedream}. To address the aforementioned issue, prior works on pipeline parallelism, including PipeDream \cite{pipedream} and GPipe \cite{gpipe}, propose pipelining across devices for better resource utilization; however, as the number of layers and devices increases, pipeline parallelism still faces the trade-off between resource utilization in synchronous parameter updates and staleness in asynchronous parameter updates \cite{pipedream}. Moreover, to fully fill the pipeline with useful computation, each device needs to store the activations at the partition boundaries for all mini-batches that enter the pipeline. Therefore, the maximum number of devices that pipeline parallelism can support is limited by the memory capacity of a single device.

Algorithmically, the fundamental reason for this scalability limitation observed from prior works is that the back-propagation (BP) algorithm \cite{backprop} imposes a \emph{strong sequential dependency} between layers during the gradient computation. Since computing systems evolve to have more and more parallel nodes \cite{dark_silicon, mcm_gpu}, in this work, we aim at exploring the following question: \emph{How can BP scale efficiently when the number of layers and devices keeps increasing into the foreseeable future?}

\begin{figure}[]
    \centering\vspace{-0.0cm}
    \includegraphics[width=\linewidth]{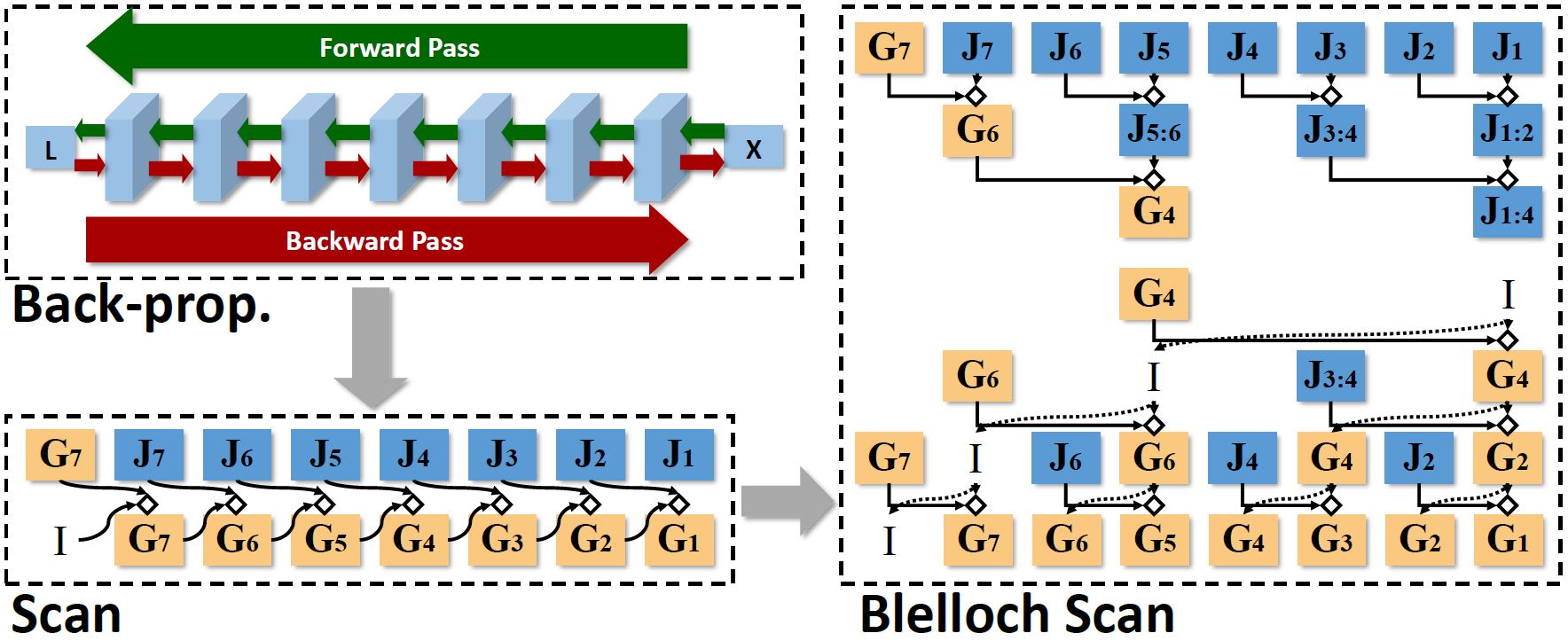}\vspace{-0.2cm}
    \caption{BP as a \emph{scan} operation, scaled by our modified version of the \emph{Blelloch scan} algorithm.}
    \label{fig:linear_to_blelloch_scan}
    \vspace{-0.6cm}
\end{figure}

To answer this question, we utilize a primitive operation called \emph{Scan} \cite{blelloch} that performs an in-order aggregation on a sequence of values and returns the partial result at each step. Parallel algorithms \cite{hillis_steele, blelloch} have been developed to scale the scan operation on massively parallel systems. We observe that BP is mathematically similar to a scan operation on the transposed Jacobian matrix \cite{jacobian} of each layer and the gradient vector of the output from the last layer. Inspired by this key observation, we \emph{restructure} the strong sequential dependency of BP, and present a new method to scale \textbf{B}ack-\textbf{p}ropagation by \textbf{P}arallel \textbf{S}can \textbf{A}lgorithm (\textbf{BPPSA}). Our major contributions are summarized below.

$\bullet$ We reformulate BP as a \emph{scan} operation and modify the \emph{Blelloch scan} algorithm \cite{blelloch} to efficiently scale BP in a parallel computing environment. Our method has a theoretical step complexity\footnote{Step complexity (detailed in Section~\ref{sec:complexity_analysis}) quantifies the runtime of a parallel algorithm.} of $\Theta (\log n)$, where $n$ represents the number of devices into which a model is partitioned, compared to $\Theta (n)$ of the na\"ive implementation of model parallelism. Moreover, our algorithm does not have the theoretical scalability limit by the memory capacity of a single device as pipeline parallelism does. As an example, Figure~\ref{fig:linear_to_blelloch_scan} shows how BP for training a network composed of 7 layers (blue cubes) can be reformulated into a scan operation on the transposed Jacobian matrices (blue squares) of this network and the final gradient vector (yellow squares), as well as how this scan operation can be scaled by BPPSA.
    
$\bullet$ Generating, storing and processing full Jacobian matrices are usually considered to be prohibitively expensive. However, we observe that the Jacobians of many types of layers (e.g., convolution, activation, max-pooling) can be extremely sparse where we can leverage sparse matrix format \cite{csr} to reduce the runtime and storage costs; more importantly, the positions of input-independent zeros in this case are deterministic, which leads to potentially more optimized implementations of sparse matrix libraries. Based on these observations, we develop routines to efficiently generate sparse transposed Jacobians for various operators.

$\bullet$ As a proof of concept, we evaluate BPPSA on a vanilla Recurrent Neural Network (RNN) \cite{elman_rnn} training with synthetic datasets, as well as a RNN with Gated Recurrent Units (GRU) \cite{gru} training with the IRMAS dataset \cite{irmas}. Our method achieves a maximum $2.75\times$ speedup in terms of the overall (end-to-end) training time, and up to $108\times$ speedup on the backward pass, compared to the baseline BP approach which under-utilizes the GPU. Moreover, we demonstrate that the retraining of pruned networks \cite{prune_2015, prune_2016, channel_pruning} (e.g., pruned VGG-11 \cite{vgg}) can also be a practical use case of BPPSA.
\vspace{-0.2cm}
\section{Background and Motivation}
\vspace{-0.1cm}
\subsection{Problem Formulation} \label{sec:formulation}

\begin{figure*}[t]
    \centering\vspace{-0.0cm}
    \includegraphics[width=0.85\linewidth]{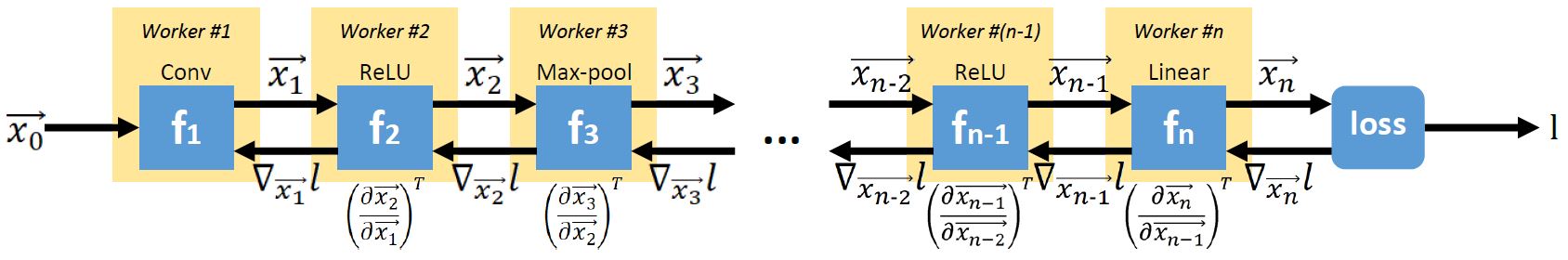}\vspace{-0.3cm}
    \caption{A visualization of the formulation in Section~\ref{sec:formulation} on convolutional neural networks. Different parts of the model can be distributed to different devices (workers).}
    \label{fig:mlp}
    \vspace{-0.5cm}
\end{figure*}

We conceptualize a deep learning model as a vector function $f$ composed of sub-functions $\vec{x}_i = f_i(\vec{x}_{i-1} \ssep \param_i)$:\vspace{-0.1cm}
\begin{equation}
f(. \ssep \param_1 , ..., \param_n) = f_1(. \ssep \param_1) \circ ... \circ f_n(. \ssep \param_n) \vspace{-0.1cm}
\end{equation}
where $\param_i, i \in \{1, ..., n\}$ are the parameters of the model. The model is evaluated by an objective function $l(f(\vec{x}_0 \ssep \param_i , i \in \{1, ..., n\}))$, where $\vec{x}_0$ is the initial input to the model. Figure~\ref{fig:mlp} visualizes a convolutional neural network conceptualized in this formulation.

To train the model $f$, a first-order optimizer requires the gradients $\nabla_{\param_i} l$, which are derived from the gradients $\nabla_{\vec{x}_i} l$:\vspace{-0.1cm}
\begin{equation} \label{eqn:update_param}
[\nabla_{\param_1} l, ..., \nabla_{\param_n} l] \gets [ (\frac{\partial \vec{x}_1}{\partial \param_1})^T \nabla_{\vec{x}_1} l , ..., (\frac{\partial \vec{x}_n}{\partial \param_n})^T \nabla_{\vec{x}_n} l  ]\vspace{-0.1cm}
\end{equation}
where $ \frac{\partial \vec{x}_i}{\partial \param_i} $ is the Jacobian matrix of the output $\vec{x}_i$ of $f_i$ to its parameters $\param_i$. To derive $\nabla_{\vec{x}_i} l$ given $\nabla_{\vec{x}_n} l$, BP \cite{backprop} solves the following recursive equation, from $i = n-1$ to $i = 1$:\vspace{-0.1cm}
\begin{equation} \label{eqn:backprop}
\nabla_{\vec{x}_i} l \gets (\frac{\partial \vec{x}_{i+1}}{\partial \vec{x}_{i}})^T \nabla_{\vec{x}_{i+1}} l, \forall i \in \{n-1, ..., 1\}\vspace{-0.1cm}
\end{equation}
where $\frac{\partial \vec{x}_{i+1}}{\partial \vec{x}_{i}}$ is the Jacobian matrix of the output $\vec{x}_{i+1}$ of $f_{i+1}$ to its input $\vec{x}_{i}$. Equation~\ref{eqn:update_param} itself does not have dependency along $i$; therefore, the computation of $\nabla_{\param_i} l$ can be parallelized if $\nabla_{\vec{x}_{i}} l$ are available. However, Equation~\ref{eqn:backprop} imposes a \emph{strong sequential dependency} along $i$ where the computation of $\nabla_{\vec{x}_{i}} l$ can not begin until the computation of $\nabla_{\vec{x}_{i+1}} l$ finishes, and therefore, hinders the scalability when multiple workers (defined as instances of execution; e.g., a core in a multi-core CPU) are available.
\vspace{-0.2cm}
\subsection{Prior Works} \label{sec:prior_works}
To increase the utilization of hardware resources in model parallelism, prior works, e.g., PipeDream \cite{pipedream} and GPipe \cite{gpipe}, propose to pipeline the computation in the forward and backward passes across devices. However, these solutions are not ``silver bullets'' to scalability due to the following reasons.

First, both PipeDream \cite{pipedream} and GPipe \cite{gpipe} require storing activations and/or multiple versions of weights for all batches that enter the pipeline. Although GPipe's re-materialization \cite{training_deep_nets_with_sublinear_mem} can mitigate memory usage, the theoretical per-device space complexity grows linearly with the length of the pipeline (i.e., the number of devices).\footnote{Appendix~\ref{append:gpipe_mem} includes a detailed space complexity analysis.} Thus, the maximum number of devices that pipeline parallelism can support is limited by the memory capacity of a single device (e.g., the GPU global memory), and such memory capacity is not expected to grow significantly in the foreseeable future \cite{memory_scaling}.

Second, if the parameter updates are partially asynchronous as proposed in PipeDream \cite{pipedream}, the resulting staleness may effect the convergence for adaptive optimizers such as Adam \cite{adam} (Appendix~\ref{append:pipedream_staleness}). If the gradient updates are fully synchronized as proposed in GPipe \cite{gpipe}, the ``bubble of idleness'' between the forward and backward passes increases linearly with the length of the pipeline, which linearly reduces the hardware utilization and leads to diminishing returns.

Our approach fundamentally differs from these key prior works \cite{pipedream, gpipe} in the following ways. First, instead of following the dependency of BP, we reformulate BP so that scaling is achieved via the Blelloch scan algorithm \cite{blelloch} which is designed for parallelism. Second, the original BP is reconstructed exactly without introducing new sources of errors (e.g., staleness); therefore, our method is agnostic to the exact first-order optimizer being used. Third, our approach becomes more advantageous as the number of devices increases, instead of diminishing returns or hitting scalability limits due to linear per-device space complexity.
\vspace{-0.0cm}
\subsection{Definition of the Scan Operation}
For a \emph{binary} and \emph{associative} operator $\oplus$ with an identity value $I$, the \emph{exclusive scan} (a.k.a., \emph{prescan}) on an input array $[a_0, a_1, a_2, ..., a_{n-1}]$ produces an output array $[I, a_0, a_0 \oplus a_1, a_0 \oplus a_1 \oplus a_2, ..., a_0 \oplus ... \oplus a_{n-2}]$ \cite{blelloch}. Parallel scan algorithms have been developed due to the importance of the scan operation and the need to leverage the computing power of emerging parallel hardware systems \cite{hillis_steele, blelloch}.  
\vspace{-0.2cm}
\section{Proposed Method: BPPSA}
\vspace{-0.1cm}
\subsection{Back-propagation as a Scan Operation} \label{sec:backprop_as_scan}

We define a \emph{binary}, \emph{associative}, and \emph{non-commutative} operator $A \fgemm B = BA$, whose identity value is the identity matrix $I$, where $A$ can be either a matrix or a vector and $B$ is a matrix. Using operator $\fgemm$, we can reformulate Equation~\ref{eqn:backprop} as calculation of the following array:\vspace{-0.1cm}
\begin{equation} \label{eqn:scan}
{\scriptsize
\begin{aligned}
[&\nabla_{\vec{x}_{n}} l, 
 \nabla_{\vec{x}_{n}} l \fgemm (\frac{\partial \vec{x}_{n}}{\partial \vec{x}_{n-1}})^T, 
 \nabla_{\vec{x}_{n}} l \fgemm (\frac{\partial \vec{x}_{n}}{\partial \vec{x}_{n-1}})^T \fgemm (\frac{\partial \vec{x}_{n-1}}{\partial \vec{x}_{n-2}})^T, \\
 &..., 
 \nabla_{\vec{x}_{n}} l \fgemm (\frac{\partial \vec{x}_{n}}{\partial \vec{x}_{n-1}})^T \fgemm ... \fgemm (\frac{\partial \vec{x}_{2}}{\partial \vec{x}_{1}})^T ]
\end{aligned}
}\vspace{-0.1cm}
\end{equation}
Equation~\ref{eqn:scan} can be interpreted as an \emph{exclusive scan} operation of $\fgemm$ on the following input array:\vspace{-0.1cm}
\begin{equation} \label{eqn:scan_input}
[ \nabla_{\vec{x}_{n}} l, (\frac{\partial \vec{x}_{n}}{\partial \vec{x}_{n-1}})^T, (\frac{\partial \vec{x}_{n-1}}{\partial \vec{x}_{n-2}})^T, ..., (\frac{\partial \vec{x}_{2}}{\partial \vec{x}_{1}})^T, (\frac{\partial \vec{x}_{1}}{\partial \vec{x}_{0}})^T  ]\vspace{-0.1cm}
\end{equation}
\vspace{-0.6cm}
\subsection{Scaling Back-propagation with the Blelloch Scan Algorithm}

\begin{figure*}[t]
    \vspace{-0.0cm}
    \centering
    \includegraphics[width=0.8\linewidth]{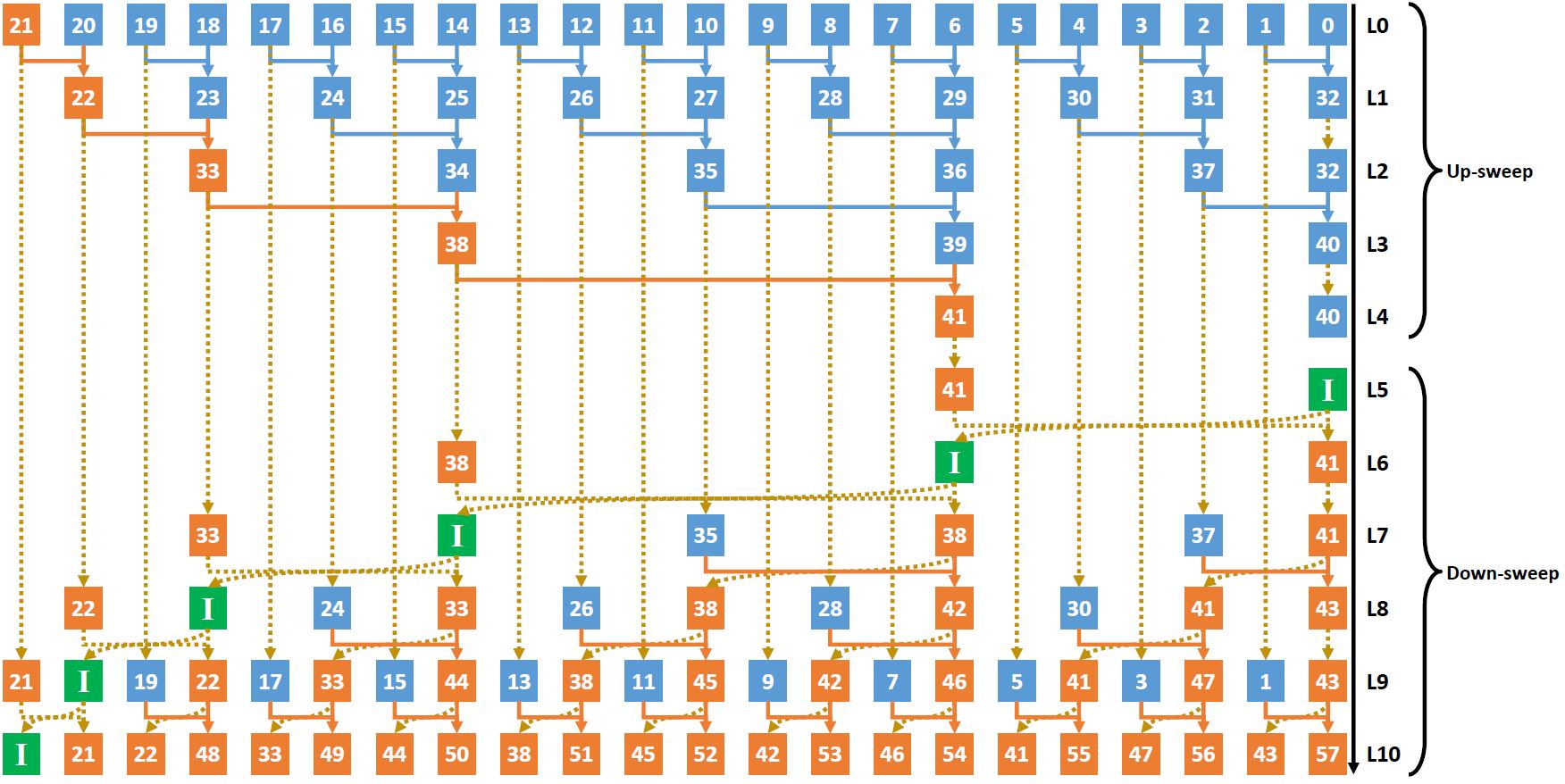}\vspace{-0.3cm}
    \caption{Applying our algorithm on the convolutional layers of VGG-11 \cite{vgg}. Blue, orange, and green squares represent transposed Jacobian matrices, gradient vectors, and \emph{symbolic} identity matrices respectively. Blue solid lines, orange solid lines, and yellow dash lines represent matrix-matrix multiplications, matrix-vector multiplications, and \emph{logical} data movements (that do not always have to be performed explicitly) respectively.}
    \label{fig:blelloch_vgg11}
    \vspace{-0.5cm}
\end{figure*}

\begin{figure}[t]
    \centering
    \begin{subfigure}[]{0.47\linewidth}
        \centering
        \includegraphics[width=0.3\linewidth]{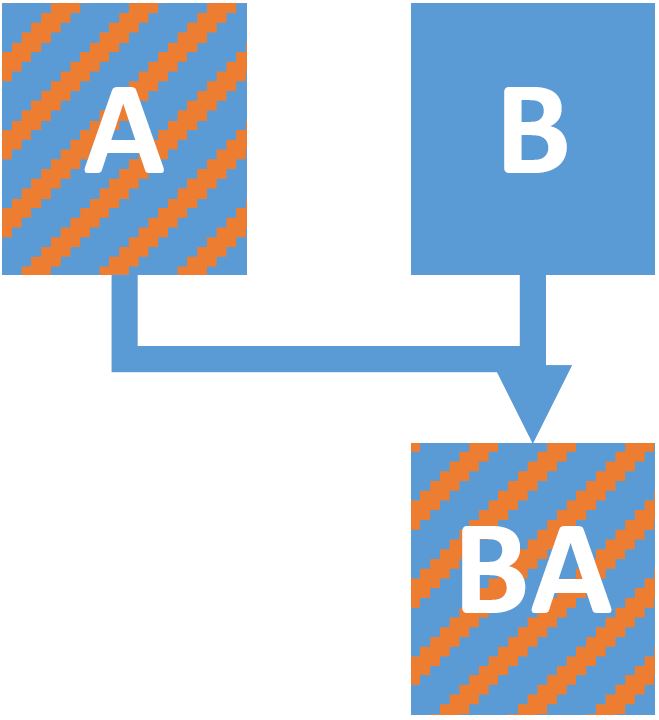}\vspace{-0.1cm}
        \caption{Up-sweep: $B \gets A \fgemm B$.}
        \label{fig:up_sweep}
    \end{subfigure}
    \hfill
    \begin{subfigure}[]{0.47\linewidth}
        \centering
        \includegraphics[width=0.3\linewidth]{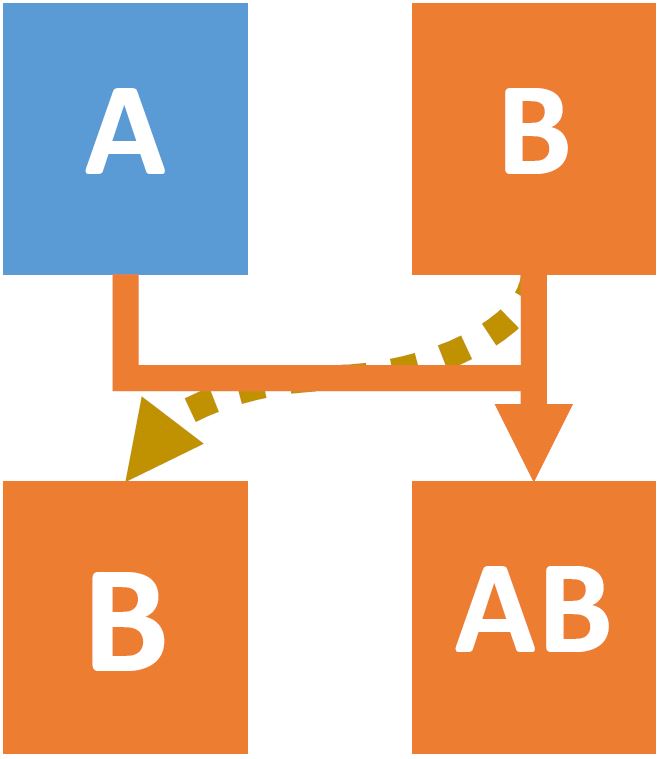}\vspace{-0.1cm}
        \caption{Down-sweep: $A, B \gets B, B \fgemm A$.}
        \label{fig:down_sweep}
    \end{subfigure}\vspace{-0.3cm}
    \caption{Visualizations of the primitive operations performed in the up-sweep and the down-sweep phases.}
    \label{fig:up_down_sweep}
    \vspace{-0.5cm}
\end{figure}

We parallelize the computation of Equation~\ref{eqn:scan} on multiple workers with the Blelloch scan algorithm \cite{blelloch}, formally described in Algorithm~\ref{algo:blelloch}. The algorithm contains two phases: \emph{up-sweep} and \emph{down-sweep}. As an example, Figure~\ref{fig:blelloch_vgg11} visualizes this algorithm applied on the convolutional layers of VGG-11 \cite{vgg} with levels L0-L4 as the up-sweep and levels L5-L10 as the down-sweep. Only the up-sweep phase contains matrix-matrix multiplications. Due to the \emph{non-commutative} property of the operator $\fgemm$, we have to reverse the order of operands for $\fgemm$ during the down-sweep phase. This modification is reflected on line 13 of Algorithm~\ref{algo:blelloch} and visualized in Figure~\ref{fig:down_sweep}.

\begin{algorithm}[]
  \scriptsize
  \caption{Modified Blelloch Scan Algorithm}\label{algo:blelloch}
  \begin{algorithmic}[1]
    \Require $a = [ \nabla_{\vec{x}_{n}} l, (\frac{\partial \vec{x}_{n}}{\partial \vec{x}_{n-1}})^T, ..., (\frac{\partial \vec{x}_{1}}{\partial \vec{x}_{0}})^T ] $ \Comment{Input array of Equation~\ref{eqn:scan_input}}
    \Ensure $a = [I, \nabla_{\vec{x}_{n}} l, ..., \nabla_{\vec{x}_{1}} l]$ \Comment{$\nabla_{\vec{x}_{i}} l$ for Equation~\ref{eqn:update_param}; computed \textbf{in-place}}
    \For{$d \gets 0$ to $\ceil{\log (n+1)} - 2$} \Comment{Up-sweep Phase}
      \FORALLP{$i \gets 0$ to $(n - 2^{d})$ by $2^{d+1}$}
        \State $(l, r) \gets (i + 2^d - 1, \min(i + 2^{d+1} - 1, n))$
        \State $a[r] \gets a[l] \fgemm a[r]$
      \ENDFAP
    \EndFor
    \State $a[n] \gets I$
    \For{$d \gets \ceil{\log (n+1)} - 1$ to $0$} \Comment{Down-sweep Phase}
      \FORALLP{$i \gets 0$ to $(n - 2^d)$ by $2^{d+1}$}
        \State $(l, r) \gets (i + 2^d - 1, \min(i + 2^{d+1} - 1, n))$
        \State $T \gets a[l]$
        \State $a[l] \gets a[r] $
        \State $a[r] \gets a[r] \fgemm T $ \Comment{\textbf{Modification:} Reverse the operands of $\fgemm$.}
      \ENDFAP
    \EndFor
  \end{algorithmic}
\end{algorithm}

\vspace{-0.2cm}
\subsection{Jacobian Matrices in Sparse Format} \label{sec:jcb_in_sparse_format}
\begin{figure}[t]
    \centering
    \begin{subfigure}{0.42\linewidth}
        \centering
        \includegraphics[width=\linewidth]{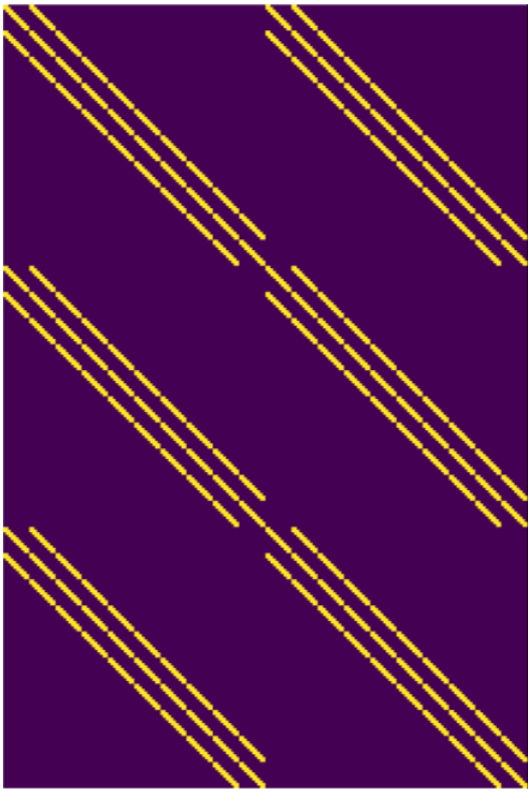}\vspace{-0.1cm}
        \caption{Convolution}
        \label{fig:conv2d_jcb}
    \end{subfigure}\hfill
    \begin{subfigure}{0.28\linewidth}
        \centering
        \includegraphics[width=0.58\linewidth]{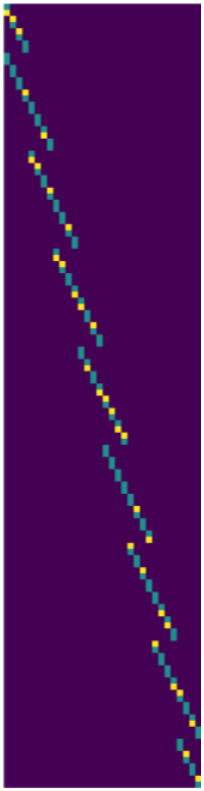}\vspace{-0.1cm}
        \caption{Max-pooling.}
        \label{fig:maxpool_jcb}
    \end{subfigure}\hfill
    \begin{subfigure}{0.3\linewidth}
        \centering
        \includegraphics[width=\linewidth]{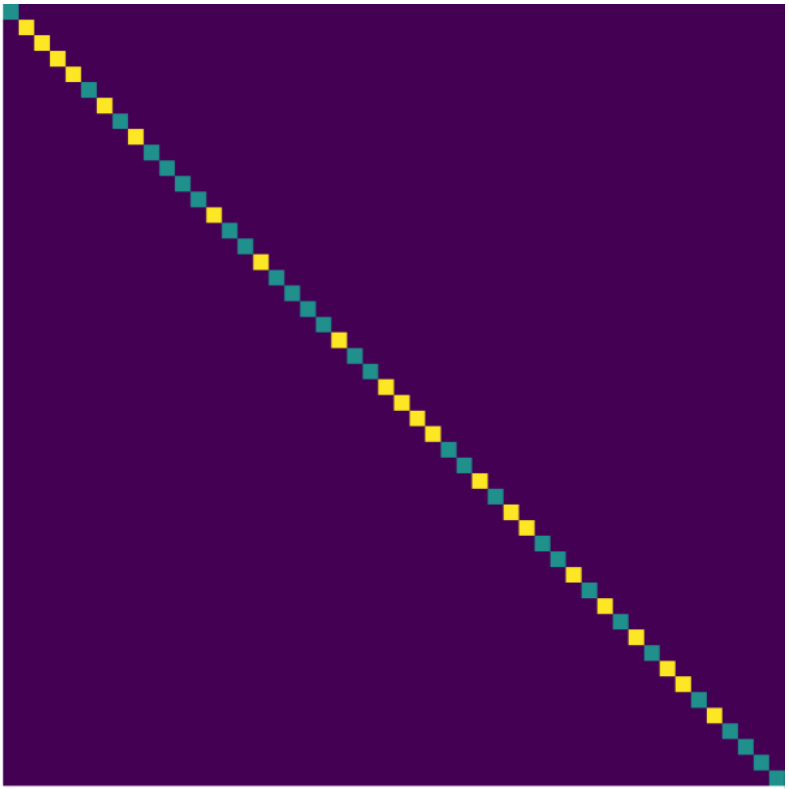}\vspace{-0.1cm}
        \caption{ReLU.}
        \label{fig:relu_jcb}
    \end{subfigure}\vspace{-0.3cm}
    \caption{Transposed Jacobians for various operators. Yellow, cyan and purple dots represents locations of non-zero elements, possible zeros and guaranteed zeros in the matrix.} \label{fig:op_jcb}
    \vspace{-0.5cm}
\end{figure}

A full Jacobian matrix $\frac{\partial \vec{x}_{i+1}}{\partial \vec{x}_{i}}$ of $f_{i+1}(. \ssep \param_{i+1})$ can be too expensive to generate, store, and process. In fact, the Jacobian of the first convolution operator in VGG-11 \cite{vgg} processing a $32 \times 32$ image occupies 768 MB of memory if stored as a full matrix, which is prohibitively large. Fortunately, Jacobian matrices of major operators (such as convolution, ReLU, and max-pooling) are usually extremely sparse as shown in Figure~\ref{fig:op_jcb}. In comparison, representing the data contained in the same Jacobian of the aforementioned convolution operator in the Compressed Sparse Row (CSR) \cite{csr} format shrinks the memory consumption down to only 6.5 MB. We can observe that there are two reasons for zeros to appear in an operator's Jacobian: \emph{guaranteed zeros} that are input ($\vec{x}_{0}$) invariant (e.g., zeros that are not on the diagonal of ReLU's Jacobian) and related to the model's architecture; and \emph{possible zeros} that depend on the input (e.g., zeros on the diagonal of ReLU's Jacobian). For any operator, the positions of guaranteed zeros (named as the \emph{sparsity pattern} for brevity) in the Jacobian is \emph{deterministic} with the model architecture and known \emph{ahead of training time}. Thus, mapping non-zero elements in the input matrices to each non-zero element in the product matrix (e.g., calculating the number of non-zeros and index merging in CSR matrix-matrix multiplication \cite{improve_perf_of_spgemm_gpu}) can be performed prior to training and removed from a generic sparse matrix multiplication routine (e.g., cuSPARSE \cite{cusparse}) to achieve significantly better performance during the training phase. As an example, the second last column of Table~\ref{tab:sparsity} shows the extremely \emph{high sparsity} of \emph{guaranteed zeros} (defined as the fraction over all elements in a matrix) for various operators in VGG-11 \cite{vgg}. In our implementation, the transposed Jacobian matrices are represented in the CSR format since it is the most straightforward and commonly used sparse matrix format; however, any other sparse matrix format can be used as an alternative, including a potentially more efficient customized sparse matrix format that utilizes the deterministic property of the current sparsity pattern, which we leave to investigate as part of our future work.

\begin{table*}[t]
\vspace{-0.2cm}
\caption{The sparsity expressions  of guaranteed zeros for various operators.} \label{tab:sparsity} \vspace{-0.2cm}
\centering
\small
\begin{threeparttable}
\begin{tabular}{lllllll}
\toprule
Operator    & Filter/Kernel Size & Input Size & Output Size & Sparsity & Examples \tnote{\textdagger} & Analytical Generation Speedup \tnote{\textsection} \\ \midrule
Convolution & $c_o \times c_i \times h_f \times w_f$            & $c_i \times h_i \times w_i$ & $c_o \times h_o \times w_o$ & \( 1-\frac{h_f w_f}{h_i w_i} \)\tnote{\textdaggerdbl} & 0.99157 & $8.3\times10^3\times$ \\[0.1cm] 
ReLU        & N/A        & $c \times h \times w$         & $c \times h \times w$ & \( 1 - \frac{1}{c h w} \) & 0.99998 & $1.2\times10^6\times$ \\[0.1cm] 
Max-pooling & $h_f \times w_f$         & $c_i \times h_i \times w_i$   & $c_o \times h_o \times w_o$ & \( 1 - \frac{h_f w_f}{c_i h_i w_i}\) & 0.99994 & $1.5\times10^5\times$ \\
\bottomrule
\end{tabular}
\begin{tablenotes}
\item[\textdagger] The examples of sparsity for the first convolution, ReLU and max-pooling operators of VGG-11 \cite{vgg} operating on $32 \times 32$ images are shown in the second last column of the table.
\item[\textdaggerdbl] Approximation when $h_i$ and $w_i$ are much greater than the padding size.
\item[\textsection] Over generating the transposed Jacobian through PyTorch's Autograd \cite{pytorch} one column at a time; measured on a Ryzen Threadripper 1950X \cite{ryzen_threadripper} machine; averaged across 1000 trials.
\end{tablenotes}
\end{threeparttable} \vspace{-0.5cm}
\end{table*}

\vspace{-0.2cm}
\subsection{Generating Jacobian Matrix in CSR Analytically} \label{sec:sp_jcb_gen}
To practically generate the Jacobian for an operator, instead of generating one column at a time either numerically \cite{calculate_jcb_numerically} or via automatic differentiation \cite{pytorch, calculate_jcb_autograd}, we develop analytical routines to generate the transposed Jacobian directly into the CSR format. Appendix~\ref{append:sparse_jcb_gen} demonstrates such analytical routines in detail for the convolution, ReLU and max-pooling operators. As proofs of concept for the potential performance benefits, the last column of Table~\ref{tab:sparsity} shows the speedup on analytical generation of the transposed Jacobians for the aforementioned operators in VGG-11 \cite{vgg}. As part of our future work to build a mature framework with automatic differentiation capability that performs training via BPPSA, we aim to provide a library that implements a ``sparse transposed Jacobian operator" (replacing the backward operator in the case of cuDNN \cite{cudnn}) for each forward operator.
\vspace{-0.2cm}
\subsection{Convergence} \label{sec:convergence}

\begin{figure}[t]
  \centering
  \begin{subfigure}{0.49\linewidth}
    \centering
    \includegraphics[width=\linewidth]{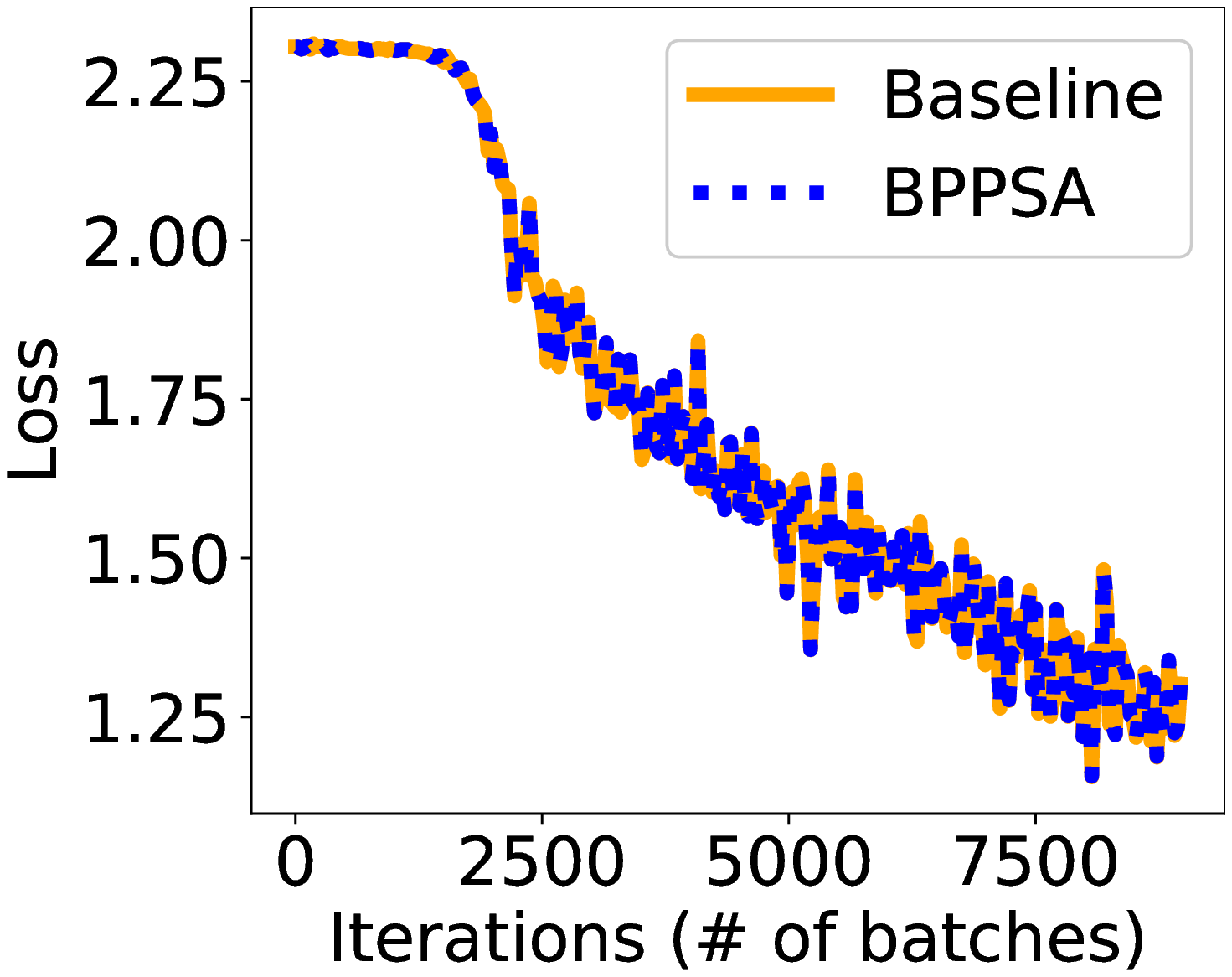}
    \caption{Training loss per iteration.}
    \label{fig:train_loss}
  \end{subfigure}\hfill
  \begin{subfigure}{0.49\linewidth}
    \centering
    \includegraphics[width=\linewidth]{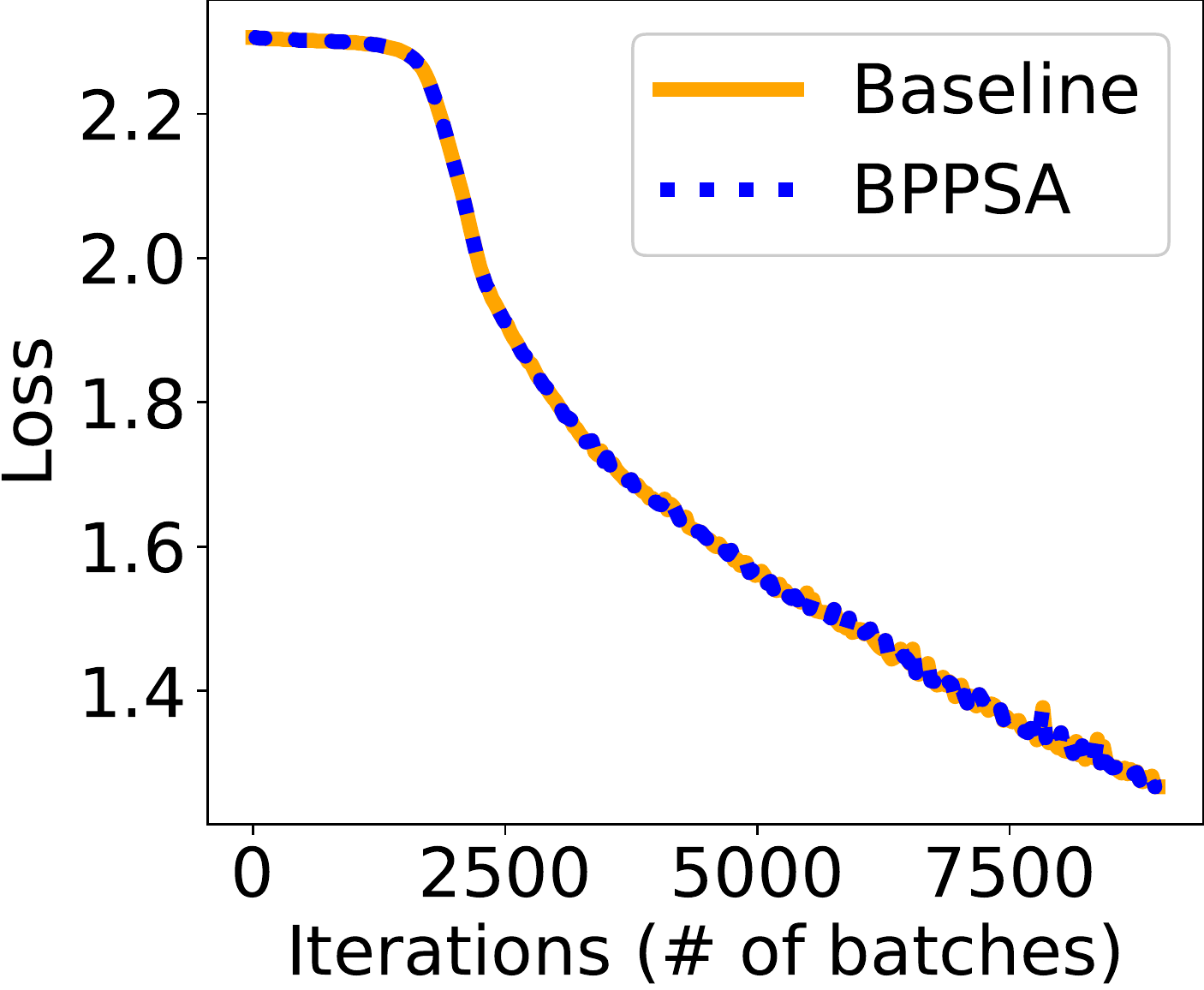}
    \caption{Test loss per iteration.}
    \label{fig:test_loss}
  \end{subfigure} \vspace{-0.3cm}
  \caption{Training and test loss per iteration for training LeNet-5 on CIFAR-10. \emph{Baseline} represents training via the PyTorch Autograd, while \emph{BPPSA} represents our method.}
  \label{fig:convergence}
  \vspace{-0.5cm}
\end{figure}

Theoretically, our algorithm is a reconstruction of BP instead of an approximation, and hence, expected to reproduce the exact same outputs. However, in practice, numerical differences could be introduced due to the change in the order of matrix multiplications. We apply our algorithm to train LeNet-5 \cite{lenet} on CIFAR-10 \cite{cifar10} to demonstrate that such numerical differences would not affect model convergence. We use a mini-batch size of 256 and the SGD \cite{sgd_with_momentum} optimizer with a learning rate of 0.001 and a momentum of 0.9. We seed the experiments with the same constant. Figure~\ref{fig:convergence} shows that the orange lines overlap with the blue lines for both training and test losses, which means our algorithm has negligible impact on the convergence compared to the original BP.
\vspace{-0.2cm}
\subsection{Complexity Analysis} \label{sec:complexity_analysis}

\paragraph{Runtime Complexity} We leverage the following definitions to quantify the complexity of a parallel algorithm: (1) \emph{step complexity} ($S$) which evaluates the minimum number of steps to finish the execution on the critical path (end-to-end) given the number of parallel workers; (2) \emph{per-step complexity} ($P$) which evaluates the runtime of a single step; and (3) \emph{work complexity} ($W$) which evaluates the number of total steps executed by all workers. For brevity, we refer to performing the scan operation serially as \emph{linear scan}, which is essentially emulating BP by using the transposed Jacobian and multiplying it with the gradient (as shown in Equation~\ref{eqn:backprop}) explicitly. Assuming the system can be conceptualized as a parallel random-access machine (PRAM) \cite{pram}, the number of workers is $p$ and the size of the input array in Equation~\ref{eqn:scan_input} is $n+1$, the step and work complexity of our algorithm can be derived as: \vspace{-0.1cm}
\begin{align}
S_{Blelloch}(n) &= \begin{cases}
\Theta(\log n) & p > n\\
\Theta(n / p + \log p) & \text{otherwise}
\end{cases}\\
W_{Blelloch}(n) &= \Theta(n)\vspace{-0.1cm}
\end{align}
compared to $S_{Linear}(n) = \Theta(n), W_{Linear}(n) = \Theta(n)$ of the linear scan (which has the same step and work complexity as BP). Therefore, in an ideal scenario where there is an unbounded number of workers with unit per-step complexity, our algorithm reduces the runtime of BP from $\bm{\Theta(n)}$ to $\bm{\Theta(\log n)}$. If, however, we consider the difference in per-step complexity between our algorithm ($P_{Blelloch}$) and the baseline ($P_{Linear}$) due to runtime difference between matrix-matrix and matrix-vector multiplications, our algorithm has a runtime of $\Theta(\log n) P_{Blelloch}$ compared to $\Theta(n) P_{Linear}$ in the baseline. There are two approaches to make our algorithm achieve a lower runtime and better scaling than the baseline. First, we can reduce $P_{Blelloch}$, which is reflected in leveraging the sparsity in the transposed Jacobian as analyzed in Section~\ref{sec:methodology_microbenchmark} and Section~\ref{sec:micro-benchmark}. Second, without lowering $P_{Blelloch}$, our algorithm can still outperform the baseline if $P_{Blelloch} / P_{Linear} < \Theta(n / \log n) $. This can occur when $n / \log n$ grows larger than the dimension of $\vec{x}_i$. The performance benefit of such case is demonstrated in Section~\ref{sec:methodology_end2end} and Section~\ref{sec:end_to_end_eval}.
\vspace{-0.4cm}
\paragraph{Space Complexity} Assuming space of storing a transposed Jacobian matrix is bounded by $M_{Jacob}$ and storing $\vec{x}_i$ is bounded by $M_{\vec{x}}$ (note that $M_{Jacob} \ll O({M_{\vec{x}}}^2)$ due to sparse matrix formats; both $M_{Jacob}$ and $M_{\vec{x}}$ are not functions of $p$), in our method, each worker requires the space of $M_{Blelloch}(n) = \Theta(max(\frac{n}{p}, 1))M_{Jacob}$ which reduces as $p$ increases until a constant $M_{Jacob}$, comparing to $M_{Pipeline} = \Theta(\frac{n}{p} + p)M_{\vec{x}}$ for pipeline parallelism which increases linearly as $p$ increases. Therefore, our method does not have the limitation of scalability on $p$, as long as each worker has the memory capacity of at least $M_{Jacob}$.
\vspace{-0.2cm}
\section{Methodology} \label{sec:method}

Although training deep learning models on thousands of devices has been proven feasible in the industry \cite{mlperf_result, resnet_in_one_hour}, setting up an experiment for such a large number of devices would require a data center of GPUs and re-implementing/optimizing our entire experiment framework, which requires both monetary and engineering resources out of reach for a typical academic research group. Thus, we set up small-scale experiments that can reflect the large-scale workloads to demonstrate the potential performance benefits of our method.

\textbf{Environment Setup}\hspace{0.2cm} Our experiments are performed on two platforms with RTX 2070 \cite{rtx2070} and RTX 2080Ti \cite{rtx2080ti} respectively (both are Turing architecture GPUs) whose specifications are listed in Table~\ref{tab:spec}\footnote{Appendix~\ref{append:v100_results} includes results on V100 \cite{v100}.}.

\begin{table}[t]
\centering
\caption{Specifications of our experiment platforms.} \label{tab:spec} \vspace{-0.2cm}
\scriptsize
\begin{tabular}{l|ll}
\toprule
GPU                     & RTX 2070               & RTX 2080Ti     \\\midrule
\begin{tabular}[c]{@{}l@{}}Number of Streaming\\ Multiprocessors (SMs)\end{tabular} & 36 & 68 \\
NVIDIA GPU Driver       & 430.50                   & 440.33.01         \\
CUDA \cite{cuda_paper}  & 10.0.130                 & 10.0.130          \\
cuDNN \cite{cudnn_paper}      & 7.5.1                    & 7.6.2             \\
PyTorch \cite{pytorch}  & 1.1.0                    & 1.2.0             \\
CPU                     & \begin{tabular}[c]{@{}l@{}}Ryzen Threadripper\\ 1950X \cite{ryzen_threadripper}\end{tabular} & \begin{tabular}[c]{@{}l@{}}EPYC \\7601 \cite{epyc_7601}\end{tabular}         \\
Host Memory             & 32GB, 2400MHz            & 128GB, 2133MHz    \\
Linux Kernel \cite{linux}            & 4.15.0-76                & 4.19.49         \\
\bottomrule
\end{tabular} \vspace{-0.4cm}
\end{table}

\textbf{Baselines}\hspace{0.2cm} We evaluate our method against \emph{PyTorch Autograd} \cite{pytorch} with cuDNN backend \cite{cudnn} which is a widely adopted and state-of-the-art implementation of BP. 

\textbf{Metrics}\hspace{0.2cm} We use three metrics to quantify the results from our evaluations: (1) \emph{wall-clock time} which measures the system-wide actual time spent on a process, (2) \emph{speedup} which is the ratio of the wall-clock time spent on the baseline over our method, and (3) \emph{FLOP} which represents the number of floating-point operations executed.

We leverage three types of benchmarks to empirically evaluate BPPSA: (1) an end-to-end benchmark of a vanilla RNN training on synthesized datasets to demonstrate the scalability benefits of BPPSA on long sequential dependency; (2) an end-to-end benchmark of a GRU training on the IRMAS dataset \cite{irmas} to demonstrate the potential of BPPSA on a more realistic workload; and (3) a micro-benchmark of a pruned VGG-11 \cite{vgg} to evaluate the feasibility of using sparse matrix format to reduce the per-step complexity of BPPSA.
\vspace{-0.2cm}
\subsection{RNN End-to-end Benchmark} \label{sec:methodology_end2end}

We set up experiments of training an RNN \cite{elman_rnn} on sequential data, which is a classical example of workloads where the runtime performance (in terms of the wall-clock time) is limited due to the \emph{strong sequential dependency}. The large number of operators $n$ is modeled through a large sequence length $T$. The large number of workers $p$ is reflected in the total number of CUDA threads that can be executed concurrently in all SMs of a single GPU, which we model through the fraction of GPU per sample (derived as one over the mini-batch size $B$).

\textbf{Datasets}\hspace{0.2cm} We synthesize the datasets $X = \{(x^{(k)}, c^{(k)})\}$ of 32000 training samples (i.e., $k \in \{0, 31999\}$) for the task of \emph{bitstream classification}. Each sample consists of a class label $c^{(k)}$ where $c^{(k)} \in \{0, ..., 9\}$ and a bitstream $x^{(k)}$ where the value $x_{t}^{(k)}$ at each time step $t \in \{0, ..., T-1\}$ is sampled from the Bernoulli distribution \cite{bernoulli, prob_textbook}:\vspace{-0.2cm}
\begin{equation}
    x_{t}^{(k)} \sim Bernoulli(0.05 + c^{(k)} \times 0.1)\vspace{-0.2cm}
\end{equation}
Equivalently, each bitstream $x^{(k)}$ can be viewed as a binomial experiment \cite{bernoulli, prob_textbook} of class $c^{(k)}$. 
The objective of this task is to classify each bitstream $x^{(k)}$ into its corresponding class $c^{(k)}$ correctly. We synthesize eight datasets with different $T$, where $T$ increases up to 30000. In reality, long sequences of input can often be found in audio signals such as speech \cite{voxceleb, spoken_wiki, chime} or music \cite{million_song, fma}. 


\textbf{Model}\hspace{0.2cm} We leverage a vanilla RNN \cite{elman_rnn} (described in Equation~\ref{eqn:rnn}) to solve the aforementioned task, since RNN is an intuitive, yet classical, deep learning model and often used to process sequential data:\vspace{-0.1cm}
\begin{equation} \label{eqn:rnn}
    \vec{h}_{t}^{(k)} = tanh(W_{ih} x_{t}^{(k)} + \vec{b}_{ih} + W_{hh} \vec{h}_{t-1}^{(k)} + \vec{b}_{hh})\vspace{-0.1cm}
\end{equation}
where $\vec{h}_{t}^{(k)}, \vec{b}_{ih}, \vec{b}_{hh} \in \mathbb{R}^{20}$. The output classes are predicted via the softmax function \cite{softmax} applied on a linear transformation to the last hidden states $\vec{h}_{T-1}^{(k)}$. The cross entropy \cite{goodfellow-deep-learning} is used as the loss function which is optimized in training via the Adam optimizer \cite{adam} with the learning rate of $1 \times 10^{-5}$. The computation of $\nabla_{\vec{h}_{t}^{(k)}} l$ during the backward pass carries the \emph{strong sequential dependency} which is the target for acceleration via BPPSA.

\textbf{Implementation}\hspace{0.2cm} We implement our modified version of the Blelloch scan algorithm as two custom CUDA kernels for the up-sweep and down-sweep phases respectively, along with a few other CUDA kernels for the preparation of the input transposed Jacobian matrices. Each level during the up-/down-sweep is associated with a separate CUDA kernel launch (in the same CUDA stream); therefore, synchronization is ensured between two consecutive levels. Each thread block is responsible for the $\fgemm$ operation (i.e. multiplication in reverse) of two matrices as well as moving the intermediate results, and the shared memory is leveraged for caching input and output matrices. Our custom CUDA kernels are integrated into the Python front-end where the RNN and the training procedure are defined through PyTorch's Custom C++ and CUDA Extensions \cite{pytorch_custom_c++_and_cuda_extensions}. For the forward pass and the baseline of PyTorch Autograd \cite{pytorch}, we simply plug in the PyTorch's \texttt{RNN} module \cite{pytorch_rnn} which calls into the cuDNN's RNN implementations (\texttt{cudnnRNNForwardTraining} and \texttt{cudnnRNNBackwardData}) \cite{cudnn}; therefore, our baseline is already much faster than implementing RNN in Python using PyTorch's \texttt{RNNCell} module \cite{pytorch_rnn} due to GEMM streaming and kernel fusions \cite{cudnn_rnn_paper}.
\vspace{-0.2cm}
\subsection{GRU End-to-end Benchmark} \label{sec:methodology_end2end_gru}
To extend the aforementioned RNN end-to-end benchmark to a more realistic setting, we evaluate the runtime performance of training a GRU \cite{gru} on the IRMAS \cite{irmas} dataset for the task of \emph{instrument classification} based on audio signals.

\textbf{Datasets}\hspace{0.2cm} We preprocess the IRMAS dataset and compute the mel-frequency cepstral coefficients (MFCC) \cite{mfcc} for each waveform audio sample via LibROSA's \cite{librosa} MFCC implementation. With different MFCC configurations as listed in Table~\ref{tab:mfcc_spec}, the preprocessing results in three sets (\emph{S}, \emph{M} and \emph{L}), reflecting the trade-off between the temporal and frequency resolutions. For all samples, we normalize the values of each coefficient across the frames to have zero mean and unit variance. We remove the first coefficient because it only represents the average power of the audio signal.

\begin{table}[]\vspace{-0.2cm}
\centering
\scriptsize
\caption{MFCC configurations and the resulting feature sizes (represented as the number of frames $F$ multiplied by the number of coefficients $C$) for the \emph{S}, \emph{M} and \emph{L} sets.} \label{tab:mfcc_spec}
\begin{tabular}{l|lll}
\toprule
Set Name & S & M & L \\ \midrule
MFCC Coefficients        & 20              & 13              & 7                \\ 
FFT Window Length        & 4096            & 2048            & 1024             \\
Hop Length               & 512             & 256             & 128              \\ \midrule
Resulting Input Features ($F \times C$) & $259 \times 38$ & $517 \times 24$ & $1034 \times 12$ \\ \bottomrule
\end{tabular} \vspace{-0.4cm}
\end{table}

\textbf{Model}\hspace{0.2cm} Since instrument classification is a more complex task than the synthetic workloads in Section~\ref{sec:methodology_end2end}, a GRU \cite{gru} (described in Equations~\ref{eqn:gru}) is used in this set of experiments.\vspace{-0.2cm}
\begin{equation}
\begin{aligned} \label{eqn:gru} 
&\vec{r}_{t} =  \sigma (W_{ir} \vec{x}_{t} + \vec{b}_{ir} + W_{hr} \vec{h}_{t-1} + \vec{b}_{hr}) \\
&\vec{z}_{t} = \sigma (W_{iz} \vec{x}_{t} + \vec{b}_{iz} + W_{hz} \vec{h}_{t-1} + \vec{b}_{hz}) \\
&\vec{n}_{t} = tanh(W_{in} \vec{x}_{t} + \vec{b}_{in} + \vec{r}_{t} \circ (W_{hn} \vec{h}_{t-1} + \vec{b}_{hn})) \\
&\vec{h}_{t} = (1 - \vec{z}_{t}) \circ \vec{n}_{t} + \vec{z}_{t} \circ \vec{h}_{t-1}
\end{aligned}\vspace{-0.2cm}
\end{equation}
where $\vec{h}_{t} \in \mathbb{R}^{20}$, $t \in \{0, ..., F - 1\}$ and $\vec{x}_{t} \in \mathbb{R}^{C}$. Since cuDNN's GRU implementation \cite{cudnn_rnn_paper} is closed source, we are unable to generate the transposed Jacobians efficiently (Appendix~\ref{append:gru_derivation}), which leads to significant \emph{overhead} in the forward pass. However, such overhead could potentially be reduced if cuDNN's source code becomes publicly available. Other settings are the same as Section~\ref{sec:methodology_end2end} with the exception of a $3 \times 10^{-4}$ learning rate.

\textbf{Implementation}\hspace{0.2cm} We directly use PyTorch's \texttt{GRU} module \cite{pytorch_rnn} which calls into the cuDNN's GRU implementations (\texttt{cudnnRNNForwardTraining} and \texttt{cudnnRNNBackwardData} with \texttt{CUDNN\_GRU}) \cite{cudnn}. We reuse the same CUDA implementation of the Blelloch scan algorithm as in Section~\ref{sec:methodology_end2end}.
\vspace{-0.2cm}
\subsection{Pruned VGG-11 Micro-benchmark} \label{sec:methodology_microbenchmark}

Despite the recent advances in network pruning algorithms \cite{prune_2015, prune_2016, channel_pruning}, there is no existing widely adopted software or hardware platform that can exploit performance benefits from pruning, as most techniques are evaluated through ``masking simulation'' which leads to the same (if not worse) runtime and memory usage. In contrast, in this work, we discover that the \emph{retraining of pruned networks} could benefit from BPPSA due to the following reason: Since the values in the Jacobian of a convolution operator only depend on the filter weights (Appendix~\ref{append:sparse_jcb_conv}), pruning the weights can lead to a higher sparsity in the Jacobian, which then reduces the per-step complexity of sparse matrix-matrix multiplications.

To evaluate the feasibility of leveraging the sparsity in the transposed Jacobian of each operator, we set up a benchmark with VGG-11 \cite{vgg}: training on CIFAR-10 \cite{cifar10}, pruning away $97\%$ of the weights in all convolution and linear operators using the technique proposed by See et al. \cite{prune_2016}, and retraining the pruned network. We choose this pruning percentage so that a similar validation accuracy is reached ($90.1\%$ v.s. $88.9\%$) after retraining for the same number of epochs ($100$) as training. We then apply BPPSA on the convolutional layers of VGG-11 to compute Equation~\ref{eqn:backprop}.

Since the sparsity pattern of the transposed Jacobian can be determined ahead of training time from the model architecture (as we show in Section~\ref{sec:jcb_in_sparse_format}), existing sparse matrix libraries which target generic cases are sub-optimal for our method. For example, cuSPARSE \cite{cusparse} calculates the number of non-zeros in the product matrix and merges the indices of the input matrices before it can perform the multiplication. Such preparations do not need to repeat across iterations in BPPSA's case and could be performed ahead of time due to the deterministic nature of the sparsity pattern. This, in turn, saves considerable amount of execution time. Therefore, due to the lack of a fair implementation, we perform the evaluation by calculating the FLOPs needed for each step in our method and the baseline implementations through \emph{static analysis}.

\vspace{-0.2cm}
\section{Evaluation} \label{sec:eval}

In this section, we present the results from the RNN end-to-end benchmark (Section~\ref{sec:methodology_end2end}), the GRU end-to-end benchmark (Section~\ref{sec:methodology_end2end_gru}) and the pruned VGG-11 micro-benchmark (Section~\ref{sec:methodology_microbenchmark}).
\vspace{-0.2cm}
\subsection{RNN End-to-end Benchmark} \label{sec:end_to_end_eval} 

\begin{figure}[]
    \vspace{-0.2cm}
    \centering
    \includegraphics[width=0.65\linewidth]{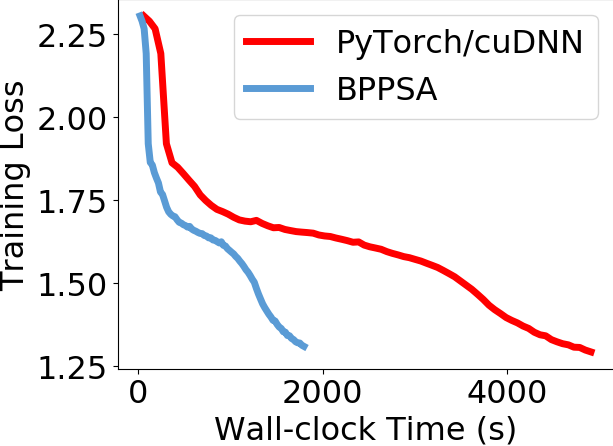}\vspace{-0.2cm}
    \caption{Training loss across wall-clock time when the RNN is trained via BPPSA (blue curve) and the PyTorch Autograd baseline with cuDNN's RNN backend (red curve).}
    \label{fig:rnn_training_loss}
    \vspace{-0.5cm}
\end{figure}

Figure~\ref{fig:rnn_training_loss} shows the training curves of loss values with respect to wall-clock time when we train the RNN for 80 epochs on the RTX 2070 GPU with the mini-batch size $B = 16$ and the sequence length $T = 1000$. This experiment can be viewed as the simplest mechanism to process sequential data such as audio signals. We observe that the blue curve (BPPSA) is roughly equivalent to the red curve (PyTorch/cuDNN baseline) scaled down by $63\%$ along the horizontal (time) axis. We conclude that, in this setting, training the RNN through BPPSA reconstructs the original BP algorithm while achieving a $2.73\times$ speedup on the overall training time and $16\times$ on the BP runtime.

\begin{figure*}[t]
     \centering
     \begin{subfigure}[]{0.32\linewidth}
         \centering
        \includegraphics[width=\linewidth]{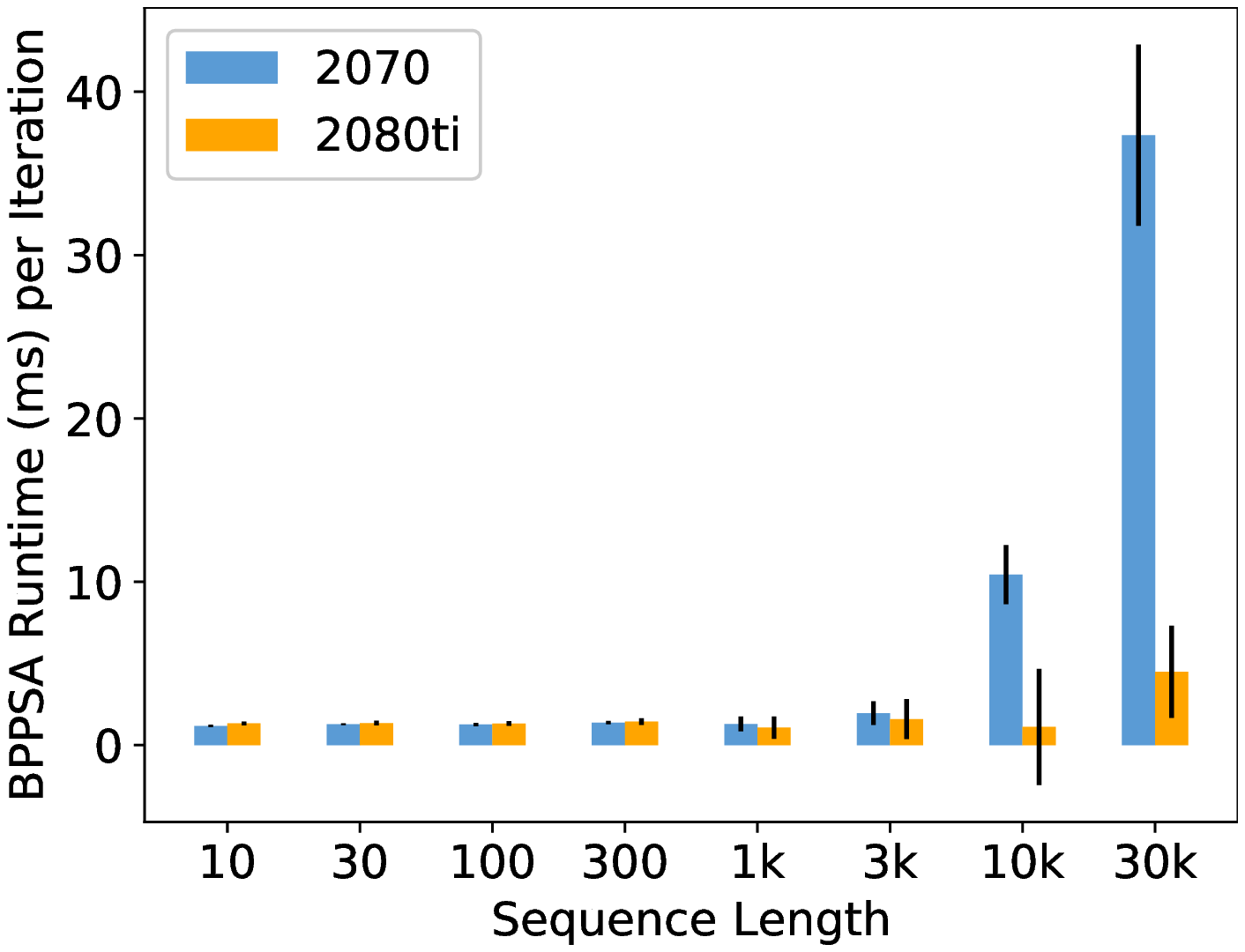}
        \caption{The BPPSA runtime per iteration as the sequence length $T$ increases.}
        \label{fig:seq_lens_to_backward_runtimes}
     \end{subfigure}
     \hfill
     \begin{subfigure}[]{0.32\linewidth}
         \centering
        \includegraphics[width=\linewidth]{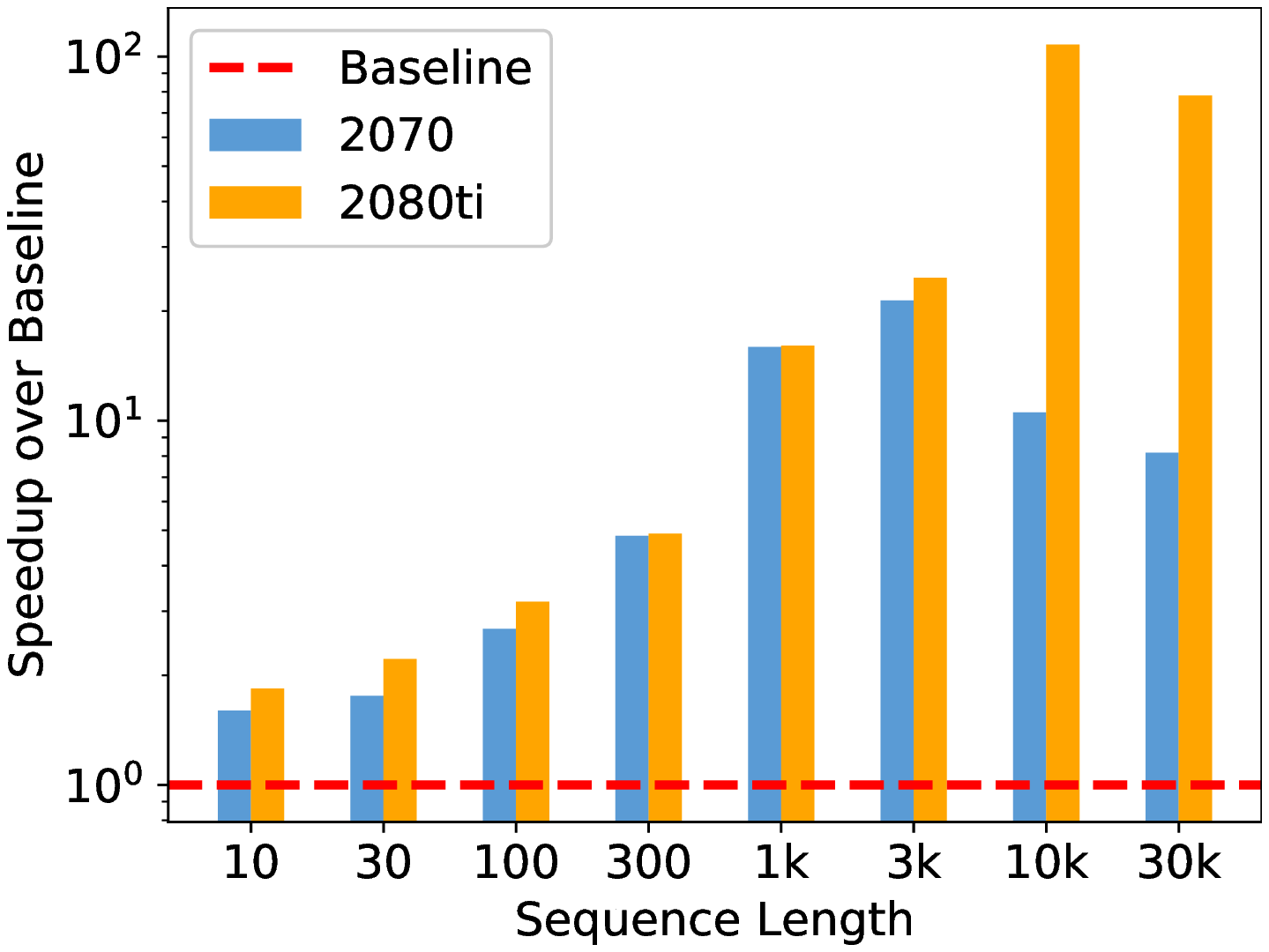}
        \caption{The backward pass speedup as the sequence length $T$ increases.}
        \label{fig:seq_lens_to_backward_speedups}
     \end{subfigure}
     \hfill
     \begin{subfigure}[]{0.32\linewidth}
         \centering
        \includegraphics[width=\linewidth]{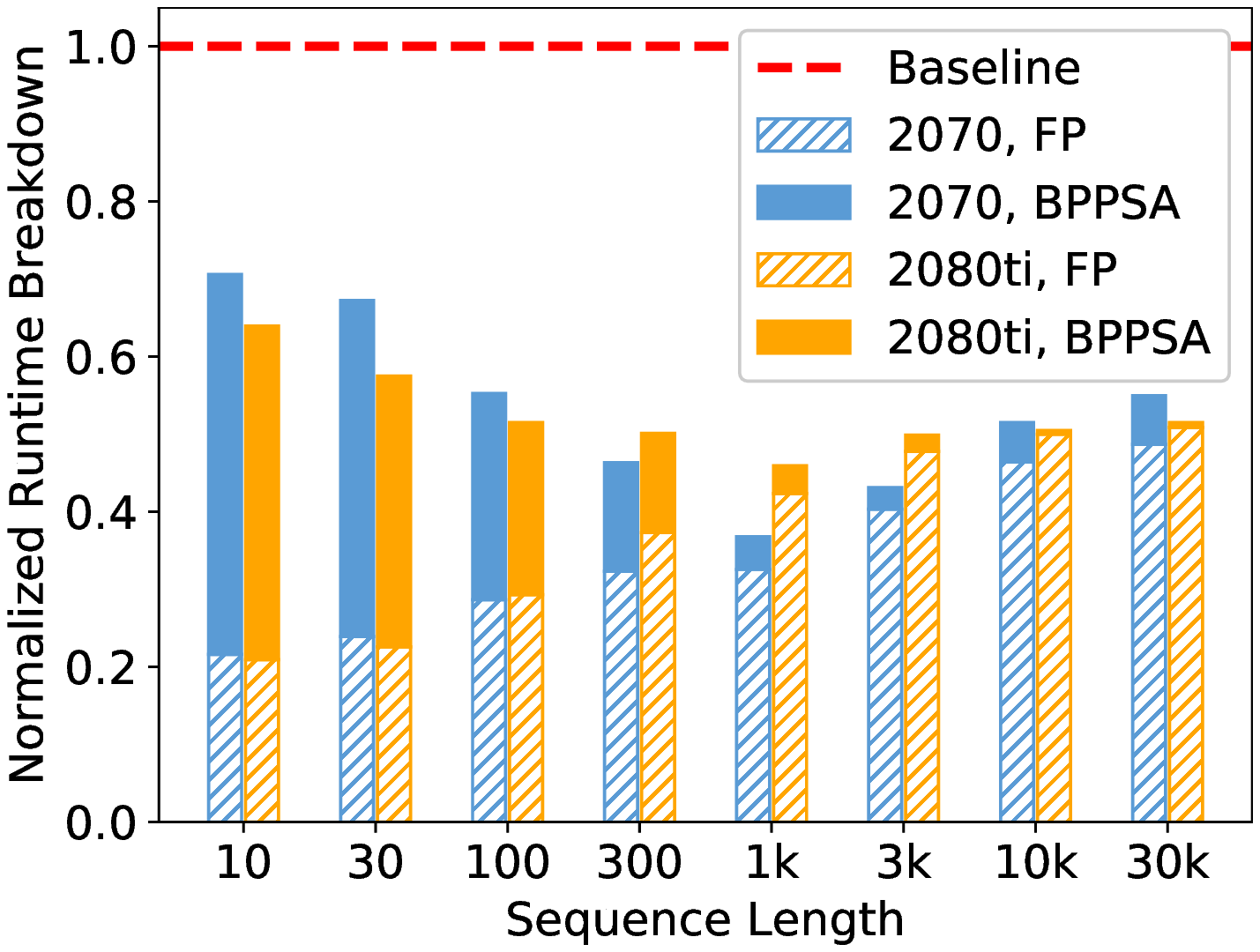}
        \caption{The runtime breakdown as the sequence length $T$ increases.}
        \label{fig:seq_lens_to_speedups}
     \end{subfigure}
     \hfill
     \begin{subfigure}[]{0.32\linewidth}
        \centering
        \includegraphics[width=\linewidth]{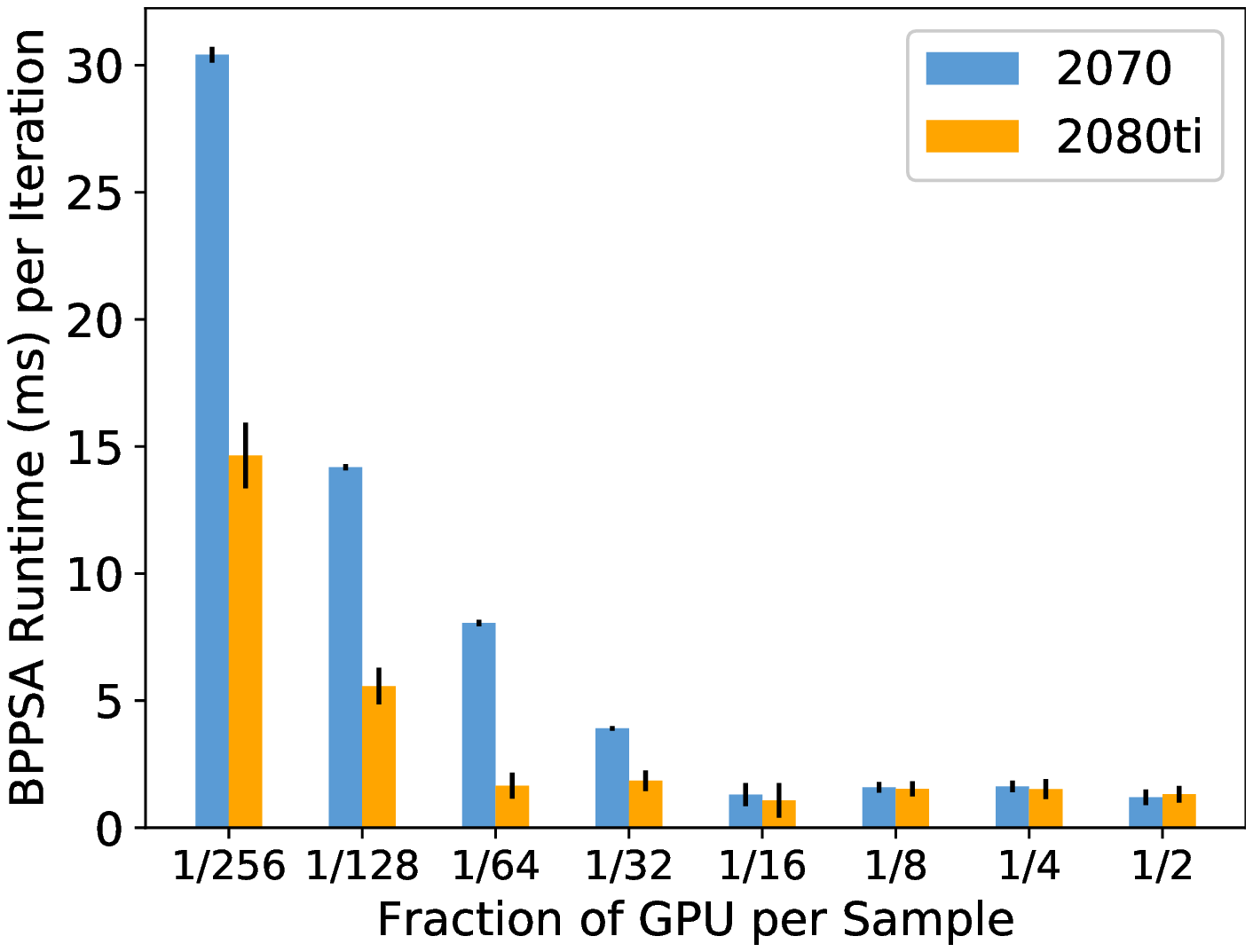}
        \caption{The BPPSA runtime per iteration as the fraction of GPU per sample ($1/B$) increases.}
        \label{fig:batch_sizes_to_backward_runtimes}
     \end{subfigure}
     \hfill
     \begin{subfigure}[]{0.32\linewidth}
        \centering
        \includegraphics[width=\linewidth]{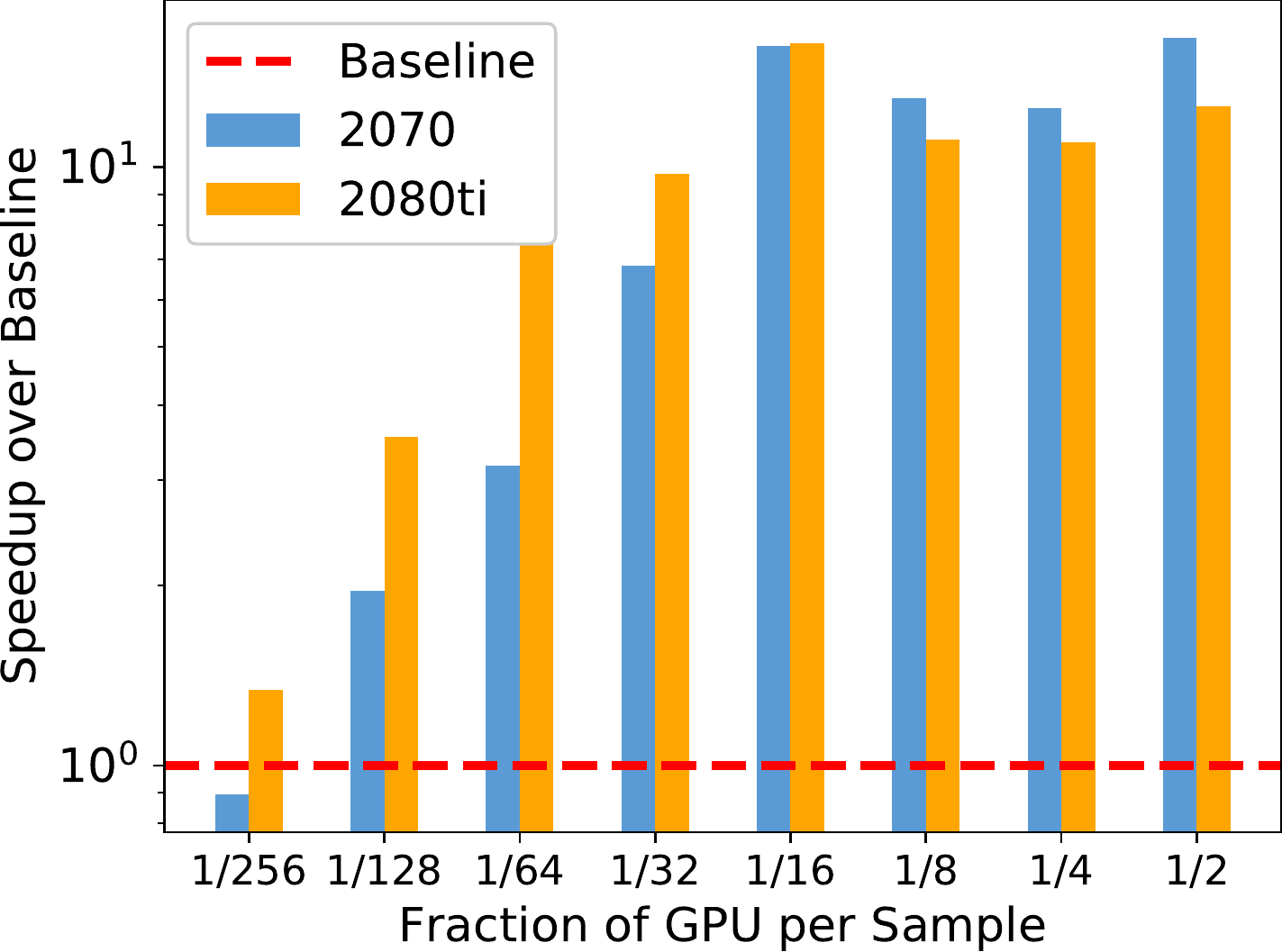}
        \caption{The backward pass speedup as the fraction of GPU per sample ($1/B$) increases.}
        \label{fig:batch_sizes_to_backward_speedups}
     \end{subfigure}
     \hfill
     \begin{subfigure}[]{0.32\linewidth}
        \centering
        \includegraphics[width=\linewidth]{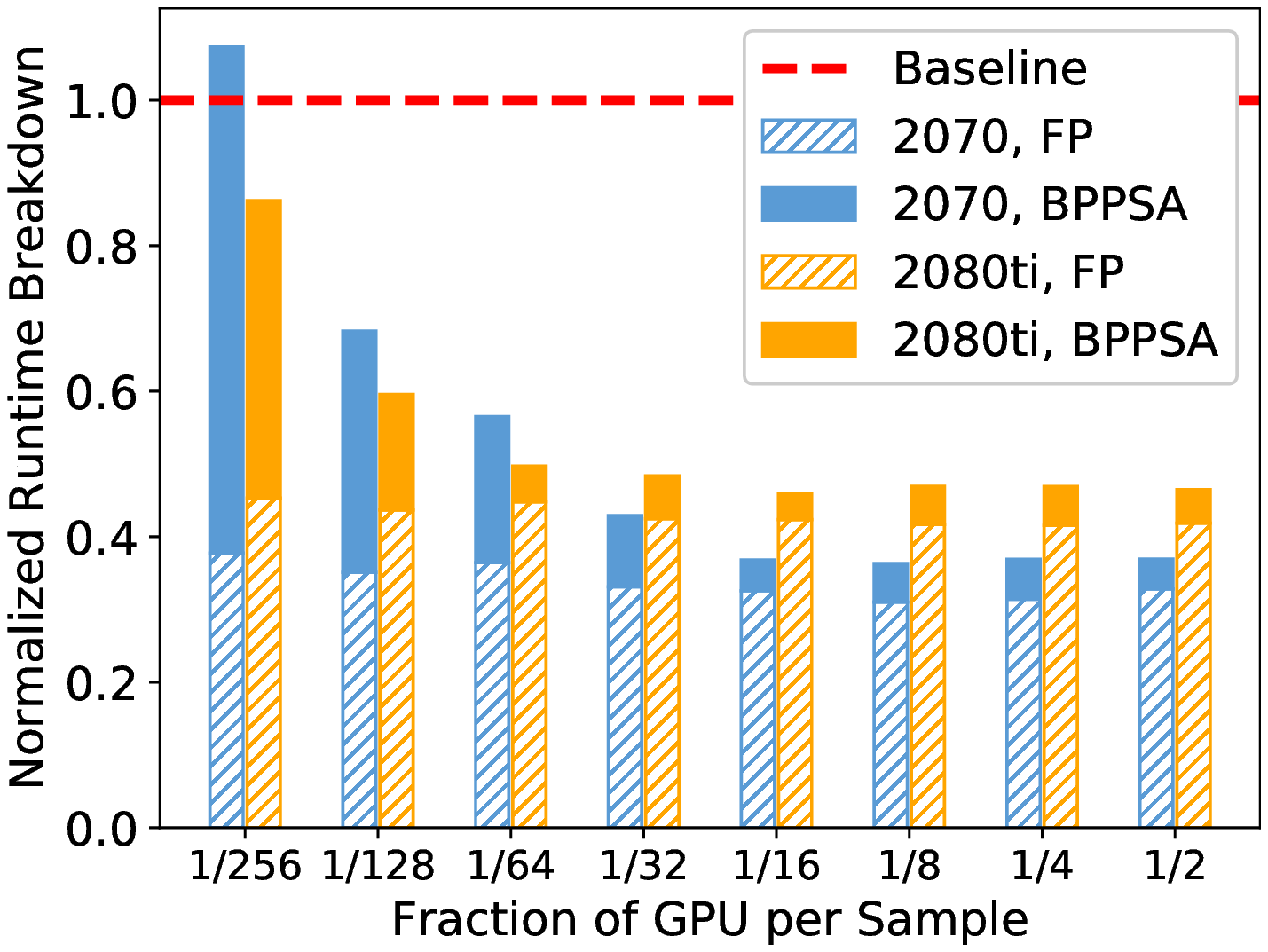}
        \caption{The runtime breakdown as the fraction of GPU per sample ($1/B$) increases.}
        \label{fig:batch_sizes_to_speedups}
     \end{subfigure}\vspace{-0.2cm}
    \caption{We report the BPPSA backward pass latency per iteration (Figure~\ref{fig:seq_lens_to_backward_runtimes} and Figure~\ref{fig:batch_sizes_to_backward_runtimes}), the backward pass speedups (Figure~\ref{fig:seq_lens_to_backward_speedups} and Figure~\ref{fig:batch_sizes_to_backward_speedups}) of BPPSA over the baseline, as well as the runtime (normalized by the baseline) breakdowns to demonstrate the overall speedups (Figure~\ref{fig:seq_lens_to_speedups} and Figure~\ref{fig:batch_sizes_to_speedups}). The fraction of GPU per sample (which reflects the number of workers $p$) is computed as one over the batch size $B$. \emph{FP} refers to the forward pass. The standard deviations of the BPPSA latency are reported as black lines in Figure~\ref{fig:seq_lens_to_backward_runtimes} and Figure~\ref{fig:batch_sizes_to_backward_runtimes}.}
    \label{fig:rnn_speedups} \vspace{-0.5cm}
\end{figure*}

\textbf{Sensitivity Analysis}\hspace{0.2cm} We measure the performance variation as the sequence length $T$ and the fraction of GPU per sample ($1/B$) vary, since those two parameters represent the total number of operators $n$ and the number of workers $p$ respectively --- key variables in the theoretical runtime of our method. To estimate the speedups, we measure the wall-clock time of training via BPPSA for a single epoch, and take the average of 20 measurements from different epochs. We then compare against training via the PyTorch/cuDNN baseline measured in the same way. We can also derive the backward pass runtime by measuring the wall-clock time of the training procedure without actually performing the backward pass, and subtracting from the total runtime (including the overhead of preparing the input transposed Jacobians).

Figure~\ref{fig:seq_lens_to_backward_runtimes}, Figure~\ref{fig:seq_lens_to_backward_speedups} and Figure~\ref{fig:seq_lens_to_speedups} show how changing the sequence length $T$ affects the backward pass and overall training time. We make three observations from these figures. First, our method scales as $n$ increases when $n$ is relatively in the same range as $p$. Second, when $n$ increases to be much larger than $p$, the performance starts to be bounded by $p$. Third, even in the range of overly large $n$, our method still achieves better utilization on massively parallel hardware than the baseline.

Figure~\ref{fig:batch_sizes_to_backward_runtimes}, Figure~\ref{fig:batch_sizes_to_backward_speedups} and Figure~\ref{fig:batch_sizes_to_speedups} show how changing the fraction of GPU per sample ($1/B$) affects the backward pass and overall training time. We can conclude that BPPSA scales as the ``effective" number of workers $p$ per sample increases (equivalently, as the batch size $B$ decreases, since the total number of SMs in the GPU is constant). In reality, determining the appropriate mini-batch size can be nontrivial: training with large batch can lead to ``generalization gap" \cite{large_mini_batch}, while training with small batch would under-utilize the hardware resources and lead to longer training time. Here, BPPSA can be viewed as offering an alternative to train with smaller mini-batch while utilizing the hardware resources more efficiently than BP.

By comparing the speedup in Figure~\ref{fig:seq_lens_to_backward_speedups} and Figure~\ref{fig:batch_sizes_to_backward_speedups} between RTX 2070 and RTX 2080Ti (RTX 2080Ti has a higher number of SMs than RTX 2070; 68 vs. 36 \cite{rtx2080ti, rtx2070}), we can observe that: (1) BPPSA achieves its maximum speedup at a higher sequence length on RTX 2080Ti than RTX 2070; (2) as the batch size $B$ increases, the speedup of BPPSA on RTX 2080Ti drops at a slower rate than RTX 2070. These two observations, together with Figure~\ref{fig:seq_lens_to_backward_runtimes} and Figure~\ref{fig:batch_sizes_to_backward_runtimes} where the BPPSA latency per iteration on RTX 2080Ti is lower than RTX 2070, are consistent with the aforementioned conclusions regarding the performance variation with the number of workers $p$. We can observe a maximum of $108\times$ backward pass speedup on RTX 2080Ti and a maximum of $2.75\times$ overall speedup on RTX 2070 (the highest backward pass speedup might not lead to the highest overall speedup due to different forward pass runtime on which BPPSA has no impact).
\vspace{-0.2cm}
\subsection{GRU End-to-end Benchmark} \label{sec:end_to_end_gru_eval}

We include the training curves of loss values with respect to the wall-clock time when we train the GRU with the preprocessed datasets in Appendix~\ref{append:gru_training_curve}. They leads to the same conclusions as the ones in Section~\ref{sec:end_to_end_eval}.

\textbf{Sensitivity Analysis}\hspace{0.2cm} To perform an analysis similar to Section~\ref{sec:end_to_end_gru_eval}, we only need to vary the batch size $B$ since the preprocessed dataset type (\emph{S}, \emph{M}, \emph{L}) already reflects the sequence length $T$. However, since the overhead of computing the transposed Jacobians during the forward pass cannot be neglected (as mentioned in Section~\ref{sec:methodology_end2end_gru}), to achieve a deeper understanding of the performance variation, we demonstrate the runtime breakdowns among the forward pass, the backward pass and the overhead. We can derive the overhead by taking the difference in the runtime of the training procedures without actually performing the backward pass between BPPSA and the PyTorch/cuDNN baseline. The measurements are averaged across 100 epochs.

\begin{figure}[t]
    \vspace{-0.0cm}
    \centering
    \includegraphics[width=\linewidth]{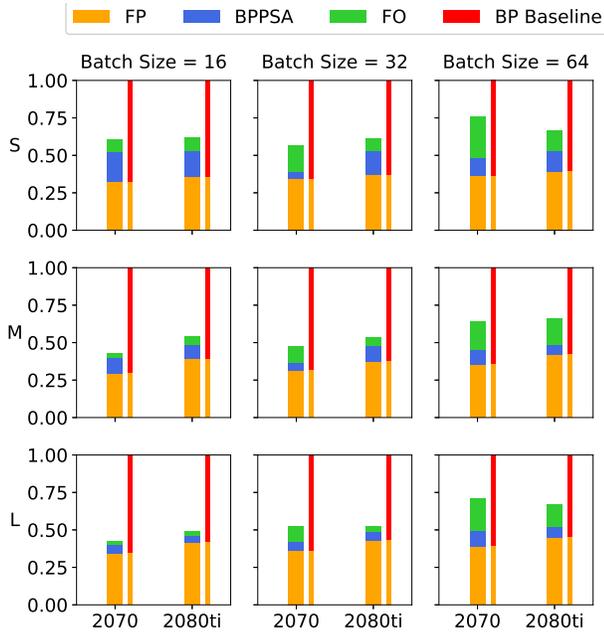}\vspace{-0.0cm}
    \caption{The runtime breakdowns in the GRU end-to-end benchmark as the dataset type (\emph{S}, \emph{M}, \emph{L}) and batch sizes $B$ vary. \emph{FP} represents the forward pass; \emph{FO} represents the forward pass overhead of computing the transposed Jacobians; \emph{BP} represents the BP baseline; and \emph{BPPSA} represents the backward pass via BPPSA. The measurements are normalized by the total runtime of the baseline (\emph{FP} + \emph{BP}).}
    \label{fig:gru_epoch_latency}
    \vspace{-0.5cm}
\end{figure}

Figure~\ref{fig:gru_epoch_latency} shows how the sequence length $T$ and batch size $B$ affect the runtimes of the forward pass, the backward pass and the overhead. We make two observations from this figure. First, our method achieves a higher speedup on the backward pass as $T$ increases (changing the preprocessed dataset from \emph{S} to \emph{L}), which reinforces the observation from Section~\ref{sec:end_to_end_eval} that our method scales well as the total number of operators $n$ increases. Second, since the maximum sequence length (1034) is not as extreme as in Section~\ref{sec:end_to_end_eval}, the backward pass runtime of BPPSA is less affected than the overhead by $B$ and the GPU model, which means $n$ is still within the same range as the number of workers $p$ in this set of experiments. The maximum overall speedup and backward pass speedup (excluding the overhead) are $2.36\times$ and $13.4\times$ respectively.

\vspace{-0.3cm}
\subsection{Pruned VGG-11 Micro-benchmark} \label{sec:micro-benchmark}

\begin{figure}
    \vspace{-0.0cm}
    \centering
    \includegraphics[width=0.8\linewidth]{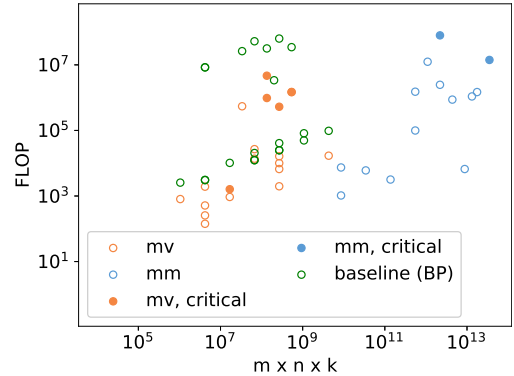}\vspace{-0.0cm}
    \caption{Measuring FLOP for each step when retraining pruned VGG-11 on CIFAR-10. \emph{mv} and \emph{mm} represent matrix-vector and matrix-matrix multiplications in BPPSA respectively. \emph{critical} indicates that the step is on the critical path. The x-axis represents the theoretical runtime complexity of the step \textbf{if} the transposed Jacobian were not encoded in a sparse format. The green circles represent the FLOP estimated for each ``gradient operator" in the BP baseline.}
    \label{fig:prune_symbolic}
    \vspace{-0.5cm}
\end{figure}

Since the sparsity of the product matrix might reduce after each multiplication, the per-step complexity might increase as the up-sweep phase progresses into deeper levels. Fortunately, we can adopt BPPSA to balance the number of levels in the up-/down-sweep phases according to the sparsity of the products on each level to achieve an overall speedup. Specifically, in this experiment, BPPSA performs the up-sweep from L0 to L2 (consistent with the notations in Figure~\ref{fig:blelloch_vgg11}), calculates the partial results that are needed for the down-sweep phase through linear scan, and then performs the down-sweep from L7 to L10.

Assuming the sparse transposed Jacobian matrices are encoded in the CSR format, Figure~\ref{fig:prune_symbolic} shows the calculated FLOP of each step in BPPSA and each ``gradient operator" in the baseline (BP) for re-training pruned VGG-11 on CIFAR-10. We observe that the green circles (baseline) have similar expected performance as the other circles (BPPSA). Thus, we can conclude that exploiting the sparsity in the transposed Jacobian is an efficient strategy that reduces the per-step complexity of our method $P_{Blelloch}$ to a level similar with the baseline $P_{Linear}$. This strategy makes the overall scalability to be ``ensured" algorithmically.
\vspace{-0.2cm}
\section{Conclusion}

In this work, we explore a novel direction to scale BP by challenging its fundamental limitation: the \emph{strong sequential dependency}. We reformulate BP into a \emph{scan} operation which is scaled by our modified version of the Blelloch scan algorithm. Our proposed algorithm, BPPSA, achieves a logarithmic, rather than linear, step complexity. In addition, BPPSA has a constant per-device space complexity; hence, its scalability is not limited by the memory capacity of each device. In our detailed evaluations, we demonstrate that performance benefits can be achieved in two important use cases. First, for the case where there is a long dependency in BP, we evaluate BPPSA by training a RNN with synthetic datasets and training a GRU with the IRMAS dataset \cite{irmas}, where our method achieves up to $2.75\times$ speedup on the overall (end-to-end) training time and $108\times$ speedup on the backward pass runtime. Second, we can reduce the per-step complexity by leveraging the sparsity in the Jacobian itself. To this end, we develop efficient routines to generate the transposed Jacobian in the CSR format, and demonstrate that the retraining of pruned networks can potentially benefit from BPPSA (as we show for a pruned VGG-11 benchmark when re-training on the CIFAR-10 dataset). We hope that our work will inspire radically new ideas and designs to improve distributed DNN training beyond the existing theoretical frameworks.

\vspace{-0.2cm}
\section*{Acknowledgements}
We want to thank Xiaodan (Serina) Tan, James Gleeson, Geoffrey Yu, Roger Grosse, Jimmy Ba, Andrew Pelegris, Bojian Zheng, Kazem Cheshmi and Maryam Mehri Dehnavi for their constructive feedback during the development of this work.
This work was supported in part by the NSERC Discovery grant, the Canada Foundation for Innovation JELF grant, the Connaught Fund, and Huawei grants.

\bibliography{main}
\bibliographystyle{sysml2019}

\section*{Summary of Appendices}

Due to space constraints, we are unable to include several important details in the main text of this paper. We provide these details in the appendix below that includes the following content:
\begin{itemize}
    \item Appendix~\ref{append:artifact} describes our open-sourced artifact and explains how to reproduce all major experiments in this work.
    \item Appendix~\ref{append:gpipe_mem} performs an analysis on the space complexity for one of the key prior works, GPipe \cite{gpipe}. Appendix~\ref{append:pipedream_staleness} describes our initial attempts to analyze PipeDream's \cite{pipedream} behavior on VGG-16 with the Adam optimizer (instead of a vanilla SGD). We use these two appendices to support our arguments in Section~\ref{sec:prior_works}.
    \item Appendix~\ref{append:sparse_jcb_gen} lists the routines that we developed to generate the transposed Jacobians for various operators directly into the CSR format. This is the complementary material for Section~\ref{sec:sp_jcb_gen}.
    \item Appendix~\ref{append:gru_derivation} shows how to derive the transposed Jacobians for GRU, and demonstrates the source of the overhead in the forward pass for our GRU end-to-end benchmark (Section~\ref{sec:methodology_end2end_gru}).
    \item Appendix~\ref{append:gru_training_curve} includes the training curves for our GRU end-to-end benchmark, which serves as a complementary material to Section~\ref{sec:end_to_end_gru_eval}.
    \item Appendix~\ref{append:v100_results} reports the additional hardware sensitivity results on the Volta-based V100 GPU \cite{v100} to validate BPPSA potential across different GPU generations.
\end{itemize}

\appendix
\section{Artifact Appendix} \label{append:artifact}

\subsection{Abstract}

We provide the source code, scripts and data that corresponds to Section~\ref{sec:method} as our artifact to reproduce the results in Section~\ref{sec:eval} and Table~\ref{tab:sparsity}. We require an x86-64 based machine with at least one NVIDIA GPU to evaluate the artifact, and NVIDIA Container Toolkit is the only software dependency to prepare. After the installation, the entire workflow (from building the needed Docker image to plotting the final results) is automated by a single \texttt{workflow.sh} script. Although the the exact numerical results produced by the artifact might vary across hardware platforms, the general trends and conclusions should be similar to the results reported in this paper.

\subsection{Artifact check-list (meta-information)}

{\small
\begin{itemize}
  \item {\bf Algorithm:} Back-propagation by Parallel Scan Algorithm (\textbf{BPPSA})
  \item {\bf Program:} RNN and GRU end-to-end benchmarks (Section~\ref{sec:methodology_end2end}, \ref{sec:methodology_end2end_gru}); a VGG-11 micro-benchmark Section~\ref{sec:methodology_microbenchmark}. All benchmarks are public, included, and automated.
  \item {\bf Compilation:} GCC 7.4.0 and CUDA 10.0 are recommended, included, and tested, although other versions of GCC and CUDA might work as well.
  \item {\bf Binary:} Scripts included to build binaries from the source code.
  \item {\bf Data set:} The synthetic datasets (Section~\ref{sec:methodology_end2end}) and IRMAS are included; approximately 4.5 GB in total.
  \item {\bf Run-time environment:} The main software dependency is NVIDIA Container Toolkit (\url{https://github.com/NVIDIA/nvidia-docker}) which dictates the OS requirements. We recommend and tested on Ubuntu 18.04.
  \item {\bf Hardware:} An x86-64 based machine with at least one NVIDIA GPU and internet access. No SUDO access needed.
  \item {\bf Run-time state:} No contentions on hardware resources (CPU, GPU, RAM, PCIe) with other processes. 
  \item {\bf Execution:} Around 57 hours in total on the RTX 2080Ti platform listed in Table~\ref{tab:spec}.
  \item {\bf Metrics:} Wall-clock time, speedup, and FLOP (Section~\ref{sec:method}).
  \item {\bf Output:} Figures that are similar to Figure~\ref{fig:rnn_training_loss}, \ref{fig:rnn_speedups}, \ref{fig:gru_epoch_latency}, \ref{fig:prune_symbolic} and \ref{fig:gru_training_loss}; Text that contains speedups similar to the last column of Table~\ref{tab:sparsity}.
  \item {\bf Experiments:} A single script is provided that automates the entire workflow.
  \item {\bf How much disk space required (approximately)?:} Approximately a total of 19.7 GB is needed.
  \item {\bf How much time is needed to prepare workflow (approximately)?:} Around one hour to install NVIDIA Container Toolkit with its dependencies.
  \item {\bf How much time is needed to complete experiments (approximately)?:} Refer to the {\bf Execution} part above.
  \item {\bf Publicly available?:} Yes.
  \item {\bf Code licenses (if publicly available)?:} MIT.
  \item {\bf Data licenses (if publicly available)?:} MIT.
  \item {\bf Workflow framework used?:} No.
  \item {\bf Archived (provide DOI)?:} 10.5281/zenodo.3605368
\end{itemize}
}
\subsection{Description}

The source code is publicly available on GitHub (\url{https://github.com/UofT-EcoSystem/BPPSA-open}) and Zenodo (\url{https://doi.org/10.5281/zenodo.3605368}). The source code and scripts only require 37.9 kB of disk space. However, the \texttt{workflow.sh} script builds a 7.7 GB Docker image, then downloads and unzips 12 GB of data.

\subsubsection{Hardware dependencies} \label{sec:artifect_hardware}

The hardware specifications used are listed in Table~\ref{tab:spec}. In general, an x86-64 based machine with at least one NVIDIA GPU and internet access is required.

\subsubsection{Software dependencies}

Although it is possible to run the experiments natively on the host machine (and, in fact, this is how our RTX 2070 platform was set up), we do not recommend this approach since installing the dependencies can be tedious, non-portable, and unsafe (due to the SUDO access requirements). Instead, we package all of the original dependencies into a Docker image which can be built natively by the \texttt{workflow.sh} script. Therefore, our artifact only requires NVIDIA Container Toolkit (\url{https://github.com/NVIDIA/nvidia-docker}). We recommend and tested on Ubuntu 18.04, however, it is possible to evaluate the artifact on other Linux distributions that NVIDIA Container Toolkit supports as well.

\subsubsection{Data sets}

The \texttt{workflow.sh} script downloads all the required datasets automatically.

\subsubsection{Models}

The RNN (Section~\ref{sec:methodology_end2end}) and GRU (Section~\ref{sec:methodology_end2end_gru}) are included. The transposed Jacobians of VGG-11 are downloaded by the \texttt{workflow.sh} script.

\subsection{Installation} \label{sec:artifect_install}

Assuming the hardware listed in Section~\ref{sec:artifect_hardware} is available, the following steps are needed to perform the installation:
\begin{enumerate}
    \item Clone the project by \texttt{git clone} \url{https://github.com/UofT-EcoSystem/BPPSA-open.git}.
    \item Install a NVIDIA GPU driver that is compatible with the GPU, the CUDA version (10.0 recommended) and the NVIDIA Container Toolkit.
    \item Install Docker Engine - Community (\url{https://docs.docker.com/install/}), then configure the \texttt{docker} group to use Docker as a non-root user. \url{https://docs.docker.com/install/linux/linux-postinstall/}).
    \item Install NVIDIA Container Toolkit (\url{https://github.com/NVIDIA/nvidia-docker}).
\end{enumerate}
We provide the \texttt{install.sh} script as a reference to the above steps 2 to 4.

\subsection{Experiment workflow}

We provide the \texttt{workflow.sh} script that automates the entire workflow consisting of the following stages:
\begin{enumerate}
    \item Build the Docker image used across experiments.
    \item Download and unzip the synthetic datasets (Section~\ref{sec:methodology_end2end}) and IRMAS.
    \item Execute the RNN (Section~\ref{sec:methodology_end2end}) and GRU (Section~\ref{sec:methodology_end2end_gru}) end-to-end benchmarks.
    \item Plot the results for the RNN and GRU end-to-end benchmarks.
    \item Evaluate the speedups for sparse transposed Jacobian generation (Section~\ref{sec:sp_jcb_gen}).
    \item Download the sparse transposed Jacobians of a regular and pruned VGG-11.
    \item Execute the VGG-11 micro-benchmark (Section~\ref{sec:methodology_microbenchmark}) and plot the results.
\end{enumerate}
After the installation in Section~\ref{sec:artifect_install}, the user only need to run the command "\texttt{./workflow.sh}" in the project root directory, which takes around 57 hours on our reference plarform with the RTX 2080Ti GPU (Table~\ref{tab:spec}).

\subsection{Evaluation and expected result}

After \texttt{./workflow.sh} finishes, a \texttt{results/} directory is created to contain the following results:
\begin{itemize}
    \item \texttt{fig\_\ref{fig:rnn_training_loss}.png} corresponding to Figure~\ref{fig:rnn_training_loss}.
    \item \texttt{fig\_\ref{fig:rnn_speedups}\_a.png}, \texttt{fig\_\ref{fig:rnn_speedups}\_b.png}, \texttt{fig\_\ref{fig:rnn_speedups}\_c.png}, \texttt{fig\_\ref{fig:rnn_speedups}\_d.png}, \texttt{fig\_\ref{fig:rnn_speedups}\_e.png} and \texttt{fig\_\ref{fig:rnn_speedups}\_f.png} corresponding to Figure~\ref{fig:rnn_speedups}.
    \item \texttt{fig\_\ref{fig:gru_epoch_latency}.png} corresponding to Figure~\ref{fig:gru_epoch_latency}.
    \item \texttt{fig\_\ref{fig:prune_symbolic}.png} corresponding to Figure~\ref{fig:prune_symbolic}.
    \item \texttt{fig\_\ref{fig:gru_training_loss}.png} corresponding to Figure~\ref{fig:gru_training_loss}.
    \item \texttt{table\_\ref{tab:sparsity}\_last\_column.txt} corresponding to the last column of Table~\ref{tab:sparsity}.
\end{itemize}
The exact numerical results might vary across hardware platforms, but the general trends should be similar to the results presented in this paper where we conducted the experiments on platforms with the RTX 2070 and 2080Ti GPUs (Table~\ref{tab:spec}). In addition, the speedups of BPPSA over BP should be easily observable in the RNN and GRU end-to-end benchmarks.

\subsection{Experiment customization}

Each stage of the workflow can be turned off independently by commenting out the corresponding lines in \texttt{workflow.sh}. The software environment can be customized by modifying \texttt{docker/Dockerfile} and rebuilding the Docker image. The parameters of the RNN and GRU end-to-end benchmarks can be customized by modifying the \texttt{code/rnn\_grid\_run.sh} and \texttt{code/gru\_grid\_run.sh} scripts which are launched by \texttt{workflow.sh} through Docker containers.

\subsection{Methodology}

Submission, reviewing and badging methodology:

\begin{itemize}
  \item \url{http://cTuning.org/ae/submission-20200102.html}
  \item \url{http://cTuning.org/ae/reviewing-20200102.html}
  \item \url{https://www.acm.org/publications/policies/artifact-review-badging}
\end{itemize}

\section{Space Complexity of GPipe} \label{append:gpipe_mem}

\begin{figure}[]
    \centering
    \includegraphics[width=\linewidth]{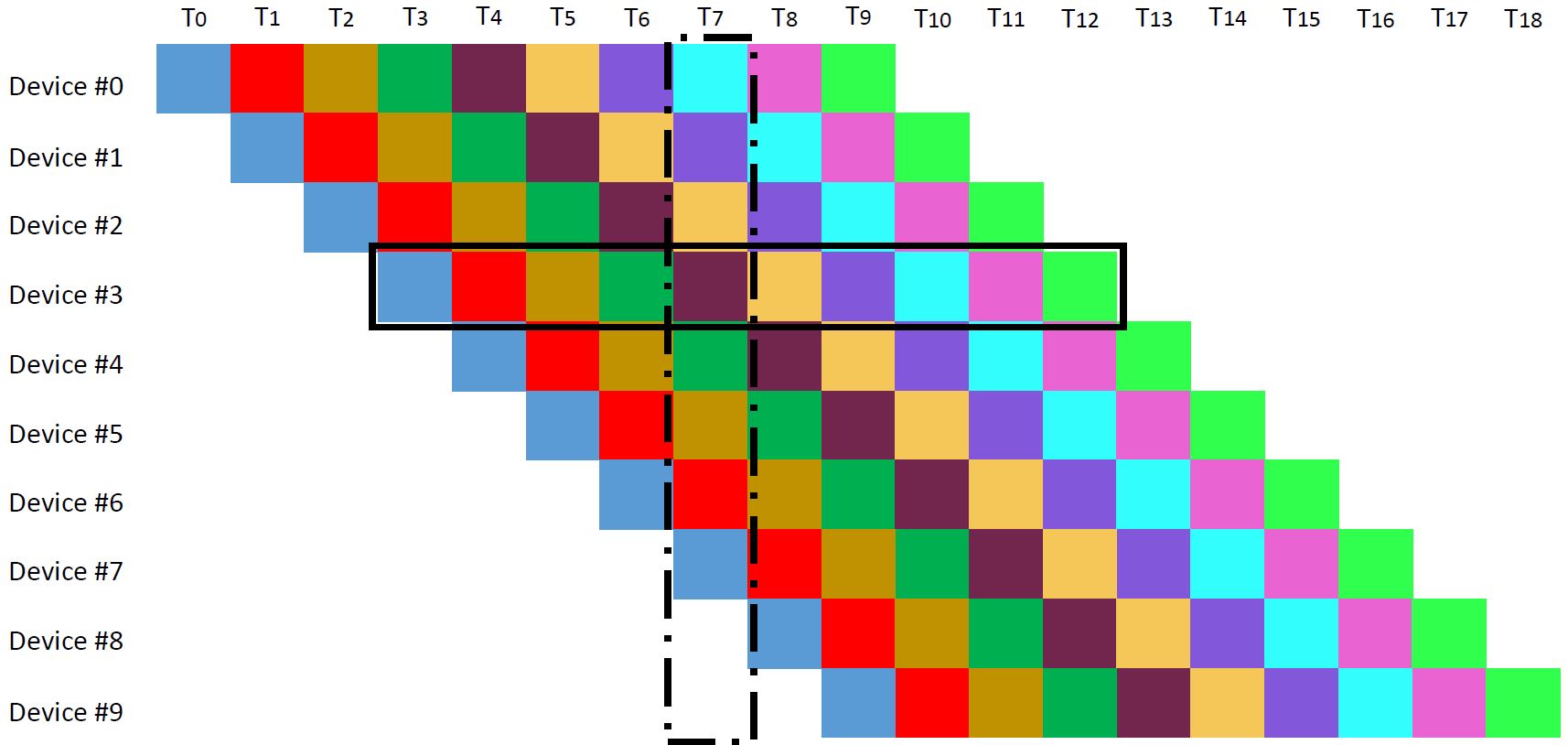}
    \caption{Timing diagram of the forward pass when distributing a model via pipeline parallelism. Each color represents an individual batch.}
    \label{fig:pipeline_parallelism}\vspace{-0.5cm}
\end{figure}

Using the notations consistent with GPipe \cite{gpipe}, with re-materialization enabled, each device reserves $\Theta(L/K)$ space for re-computing the intermediate activations of each sample in a ``micro-batch", where $L$ and $K$ are the length of the network and the number of devices in the pipeline correspondingly. As we show in Figure~\ref{fig:pipeline_parallelism}, to fully fill the pipeline with useful computation, the number of ``micro-batches" entering the pipeline (the solid black box) should be equal to the length of the pipeline (the dashed black box); thus, each device needs to store at least $\Theta(K)$ activations at the partition boundary for each sample, resulting in a $\Theta(L/K + K)$ per-device space complexity.

\section{Affect of PipeDream's Staleness on Adam} \label{append:pipedream_staleness}

\begin{figure}[t]
    \vspace{-0.0cm}
    \centering
    \includegraphics[width=\linewidth]{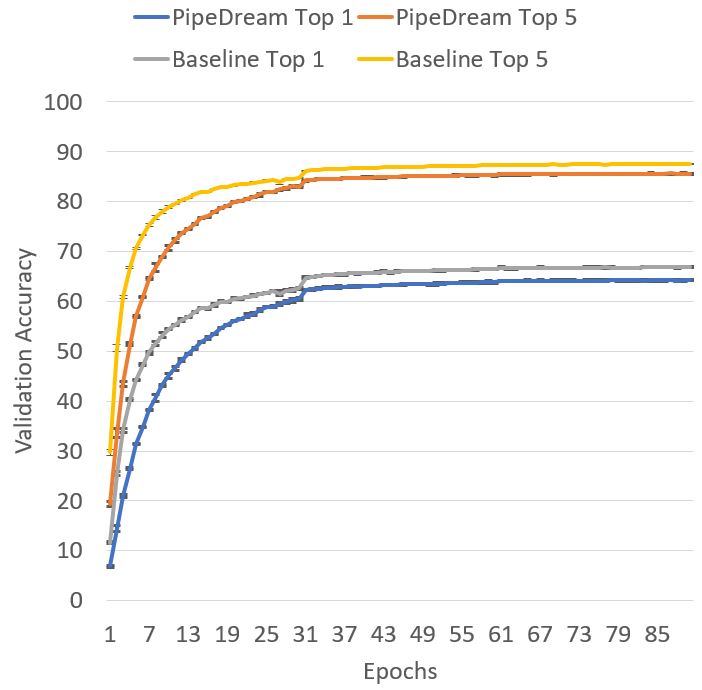}
    \caption{Top-1 and Top-5 validation accuracy across epochs for both PipeDream and the baseline. We report both the mean (as curves) and the range (as error bars). }
    \label{fig:pipedream_imagenet}
    \vspace{-0.5cm}
\end{figure}

Using the source code from \url{https://github.com/msr-fiddle/pipedream}, we reproduce PipeDream's results on VGG-16 with the same settings except the following:
\begin{itemize}
    \item 4 RTX 2080Ti GPUs (instead of 16 V100 GPUs).
    \item Mini-batch size of 32 (instead of 64).
    \item Adam optimizer with the learning rate of 0.00003 and zero weight decay (instead of SGD with the learning rate of 0.01 and the weight decay of 0.0005).
    \item 90 epochs in total (instead of 60).
    \item Step decay learning rate schedule (instead of polynomial decay). 
\end{itemize}
For the baseline, we use the source code from \url{https://github.com/pytorch/examples/tree/master/imagenet} (which is a plain VGG implementation used as one of PyTorch's official examples for ImageNet \cite{image_net}) and the same aforementioned settings except using one GPU (instead of four). We choose these settings for the following purpose: (1) to fit in the hardware resources available to us; (2) to match with a widely adopted baseline; and (3) to use the Adam optimizer instead of SGD. We run both experiments three times and record the Top-1 and Top-5 validation accuracy across epochs. We present our results in Figure~\ref{fig:pipedream_imagenet}.

We observe a 2.6\% top-1 and 1.9\% top-5 accuracy loss from PipeDream compared to the baseline. Combining with the observation that the error bars are negligible (i.e., negligible variance across runs), we conclude that at least in this case, PipeDream is not fully equivalent to the baseline for some adaptive optimizers (e.g., Adam), which differs from the optimizer-oblivious property of our work. It is important to emphasize that we \textbf{do not} imply that PipeDream will always have a negative impact on the convergence. A deeper analysis is needed on a much greater space of hyper-parameters to understand how general is such an effect on convergence with PipeDream that is beyond the scope of our work.

\section{Sparse Jacobian Generation Routines} \label{append:sparse_jcb_gen}

\subsection{Convolution} \label{append:sparse_jcb_conv}
Algorithm~\ref{algo:conv_jcb_indptr}, Algorithm~\ref{algo:conv_jcb_indices} and Algorithm~\ref{algo:conv_jcb_data} show how to generate the CSR \texttt{indptr}, \texttt{indices} and \texttt{data} arrays \cite{scipy} respectively for the transposed Jacobian of a convolution operator that has a $3 \times 3$ filter and padding size of 1.\footnote{Although we assume a specific configuration of the convolution operator here, deriving a generic routine is doable.}

\begin{algorithm}[h]
    \scriptsize
    \caption{Compute the CSR \texttt{indptr} array for the transposed Jacobian of a $3 \times 3$ convolution.} \label{algo:conv_jcb_indptr}
    \begin{algorithmic}[1]
        \Require input channels $c_{i}$, output channels $c_{o}$, input height $h_{i}$, input width $w_{i}$
        \Ensure $\texttt{indptr} \gets \texttt{malloc}(c_i h_i w_i +1)$
            \FORALLP{$i \gets 0$ to $(c_{i} h_{i} w_{i})$}
                \State $a \gets \floor{i / (h_{i}w_{i})}$
                \State $b \gets i \bmod (h_{i}w_{i})$
                \If{$b \leqslant w_{i}$}
                    \State $\texttt{indptr}[i] \gets ac_{o}(3w_{i}(3h_{i}-2))+6c_{o}b$
                \ElsIf{$b \leqslant w_{i}(h_{i}-1)$}
                    \State $\texttt{indptr}[i] \gets ac_{o}(3w_{i}(3h_{i}-2))+6c_{o}w_{i}+9c_{o}(b-w_{i})$
                \Else
                    \State $\begin{aligned}\texttt{indptr}[i] \gets& ac_{o}(3w_{i}(3h_{i}-2))+6c_{o}w_{i}+9c_{o}(w_{i}(h_{i}-2))\\&+6c_{o}(b-w_{i}(h_{i}-1))\end{aligned}$
                \EndIf
            \ENDFAP
    \end{algorithmic}
\end{algorithm}

\begin{algorithm}[h]
    \scriptsize
    \caption{Compute the CSR \texttt{indices} array for the transposed Jacobian of a $3 \times 3$ convolution.} \label{algo:conv_jcb_indices}
    \begin{algorithmic}[1]
        \Require input channels $c_{i}$, output channels $c_{o}$, input height $h_{i}$, input width $w_{i}$, \texttt{indptr} computed from Algorithm~\ref{algo:conv_jcb_indptr}
        \Ensure $\texttt{indices} \gets \texttt{malloc}(3w_{i}(3h_{i}-2) c_i c_o)$
            \FORALLP{$i \gets 0$ to $(c_i h_i w_i - 1)$}
                \State $r \gets i \bmod (h_{i}w_{i})$
                \State $\texttt{base} \gets \texttt{malloc}(9 c_o)$
                \FORALLP{$j \gets 0$ to $(c_o - 1)$}
                    \FORALLP{$k \gets 0$ to $2$}
                        \State $\begin{aligned}\texttt{base}&[9j+3k : 9j+3(k+1)] \gets (\\&[-1, 0, 1] + (j h_i + k - 1) w_i + r) \bmod (c_{o}h_{i}w_{i})\end{aligned}$
                    \ENDFAP
                \ENDFAP
                \If{$r < w_{i}$ or $r \geqslant w_{i}(h_{i}-1)$}
                    \State $\texttt{row} \gets \texttt{malloc}(6c_{o})$
                    \State $(\texttt{left}, \texttt{right}) \gets (3, 9)$ if $r < w_{i}$; $(0, 6)$ otherwise
                    \FORALLP{$j \gets 0$ to $(c_o - 1)$}
                        \State $\texttt{row}[6j : 6j + 6] \gets \texttt{base}[9j + \texttt{left}: 9j + \texttt{right}]$
                    \ENDFAP
                \Else
                    \State $\texttt{row} \gets \texttt{base}$
                \EndIf
                \State $\texttt{indices}[\texttt{indptr}[i] : \texttt{indptr}[i+1]] \gets \texttt{sorted}(\texttt{row})$
            \ENDFAP
    \end{algorithmic}
\end{algorithm}

\begin{algorithm}[h]
    \scriptsize
    \caption{Compute the CSR \texttt{data} array for the transposed Jacobian of a $3 \times 3$ convolution.} \label{algo:conv_jcb_data}
    \begin{algorithmic}[1]
        \Require input channels $c_{i}$, output channels $c_{o}$, input height $h_{i}$, input width $w_{i}$, filter \texttt{weights}, \texttt{indptr} computed from Algorithm~\ref{algo:conv_jcb_indptr}
        \Ensure $\texttt{data} \gets \texttt{malloc}(3w_{i}(3h_{i}-2) c_i c_o)$
            \FORALLP{$i \gets 0$ to $(c_i h_i w_i - 1)$}
                \State $r \gets i \bmod (h_{i} w_{i})$
                \State $m \gets \floor{i / (h_{i}w_{i})}$
                \State $\begin{aligned}\texttt{range} \gets& (\texttt{1::-1}) \text{ if } (r < w_{i})\text{;}\\& (\texttt{2:0:-1}) \text{ if } (r \geqslant w_{i}(h_{i}-1))\text{;}\\&(\texttt{2::-1}) \text{ otherwise}\end{aligned}$
                \State $\begin{aligned}\texttt{data}&[\texttt{indptr}[i]:\texttt{indptr}[i+1]] \gets \texttt{flatten}(\\&\texttt{weights}[:, m, \texttt{range}, \texttt{::-1}])\end{aligned}$
                \State Fix corner cases when $(i \bmod w_{i}) = 0$ or $(i \bmod w_{i}) = (w_{i} - 1)$.
            \ENDFAP
    \end{algorithmic}
\end{algorithm}

\subsection{ReLU}
Our methods of generating the CSR \texttt{indptr}, \texttt{indices} and \texttt{data} arrays \cite{scipy} for the transposed Jacobian of a ReLU operator are formally described in Algorithm~\ref{algo:relu_jcb_indptr}, Algorithm~\ref{algo:relu_jcb_indices} and Algorithm~\ref{algo:relu_jcb_data} respectively.

\begin{algorithm}[]
  \scriptsize
  \caption{Compute the CSR \texttt{indptr} array for the transposed Jacobian of ReLU.} \label{algo:relu_jcb_indptr}
  \begin{algorithmic}[1]
    \Require size $d$ of the (flattened) input tensor $x$
    \Ensure $\texttt{indptr} \gets \texttt{malloc}(d + 1)$
      \FORALLP{$i \gets 0$ to $d$}
        \State $\texttt{indptr}[i] \gets i$
      \ENDFAP
  \end{algorithmic}
\end{algorithm}

\begin{algorithm}[]
  \scriptsize
  \caption{Compute the CSR \texttt{indices} array for the transposed Jacobian of ReLU.} \label{algo:relu_jcb_indices}
  \begin{algorithmic}[1]
    \Require size $d$ of the (flattened) input tensor $x$
    \Ensure $\texttt{indices} \gets \texttt{malloc}(d)$
      \FORALLP{$i \gets 0$ to $(d - 1)$}
        \State $\texttt{indices} \gets i$
      \ENDFAP
  \end{algorithmic}
\end{algorithm}

\begin{algorithm}[]
  \scriptsize
  \caption{Compute the CSR \texttt{data} array for the transposed Jacobian of ReLU.} \label{algo:relu_jcb_data}
  \begin{algorithmic}[1]
    \Require the (flattened) input tensor $x$, and its size $d$
    \Ensure $\texttt{data} \gets \texttt{malloc}(d)$
      \FORALLP{$i \gets 0$ to $(d - 1)$}
        \If{$x[i] > 0$}
          \State $\texttt{data}[i] \gets 1$
        \Else
          \State $\texttt{data}[i] \gets 0$
        \EndIf
      \ENDFAP
  \end{algorithmic}
\end{algorithm}

\subsection{Max-pooling}
Assuming the stride size and the window size are the same, and we can access a tensor (named as \texttt{pool\_indices} for brevity) which specifies the indices of the elements in the input tensor that are ``pooled'' for the output tensor (documented in \cite{pytorch_maxpool2d}), our methods of generating the CSR \texttt{indptr}, \texttt{indices} and \texttt{data} arrays \cite{scipy} are formally described in Algorithm~\ref{algo:maxpool_jcb_indptr}, Algorithm~\ref{algo:maxpool_jcb_indices} and Algorithm~\ref{algo:maxpool_jcb_data} respectively.

\begin{algorithm}[]
  \scriptsize
  \caption{Compute the CSR \texttt{indptr} array for the transposed Jacobian of max-pooling.} \label{algo:maxpool_jcb_indptr}
  \begin{algorithmic}[1]
    \Require \texttt{pool\_indices}, input height $h_i$, input width $w_i$, output height $h_o$, output width $w_o$, output channels $c_o$
    \Ensure $\texttt{indptr} \gets \texttt{malloc}(c_o h_i w_i + 1)$, $\texttt{mapping} \gets \texttt{malloc}(c_o h_i w_i)$
      \FORALLP{$i \gets 0$ to $c_o h_i w_i - 1$}
        \State $\texttt{mapping}[i] \gets -1$
      \ENDFAP
      \FORALLP{$c \gets 0$ to $c_o - 1$}
        \FORALLP{$h \gets 0$ to $h_o - 1$}
          \FORALLP{$w \gets 0$ to $w_o - 1$}
            \State $i \gets c h_i w_i + \texttt{pool\_indices}[c, h, w]$
            \State $j \gets (c h_o + h) w_o + w$
            \State $\texttt{mapping}[i] \gets j$
          \ENDFAP
        \ENDFAP
      \ENDFAP
      \State $\texttt{ptr} \gets 0$
      \For{$i \gets 0$ to $(c_o h_i w_i - 1)$}
        \State $\texttt{indptr}[i] \gets \texttt{ptr}$
        \If{$\texttt{mapping}[i] \neq -1$}
          \State $\texttt{ptr} \gets \texttt{ptr} + 1$
        \EndIf
      \EndFor
      \State $\texttt{indptr}[-1] \gets \texttt{ptr}$
  \end{algorithmic}
\end{algorithm}

\begin{algorithm}[]
  \scriptsize
  \caption{Compute the CSR \texttt{indices} array for the transposed Jacobian of max-pooling.} \label{algo:maxpool_jcb_indices}
  \begin{algorithmic}[1]
    \Require \texttt{mapping} computed from Algorithm~\ref{algo:maxpool_jcb_indptr}, input height $h_i$, input width $w_i$, output height $h_o$, output width $w_o$, output channels $c_o$
    \Ensure $\texttt{indices} \gets \texttt{malloc}(c_o h_o w_o)$
      \State $\texttt{indices\_ptr} \gets 0$
      \For{$i \gets 0$ to $(c_o h_i w_i - 1)$}
        \If{$\texttt{mapping}[i] = -1$}
            \State \textbf{continue}
        \EndIf
        \State $\texttt{indices}[\texttt{indices\_ptr}] \gets \texttt{mapping}[i]$
        \State $\texttt{indices\_ptr} \gets \texttt{indices\_ptr} + 1$
      \EndFor
  \end{algorithmic}
\end{algorithm}

\begin{algorithm}[]
  \scriptsize
  \caption{Compute the CSR \texttt{data} array for the transposed Jacobian of max-pooling.} \label{algo:maxpool_jcb_data}
  \begin{algorithmic}[1]
    \Require output channels $c_o$, output height $h_o$, output width $w_o$
    \Ensure $\texttt{data} \gets \texttt{malloc}(c_o h_o w_o)$
      \FORALLP{$i \gets 0$ to $(c_o h_o w_o - 1)$}
        \State $\texttt{data} \gets 1$
      \ENDFAP
  \end{algorithmic}
\end{algorithm}

\section{Overhead Analysis of the GRU End-to-end Benchmark} \label{append:gru_derivation}

We can rewrite the GPU in Equation~\ref{eqn:gru} into the following form:
\begin{equation*}
{
\begin{aligned} \label{eqn:gru_rewrite} 
&\vec{R}_{t} =  W_{ir} \vec{x}_{t} + \vec{b}_{ir} + W_{hr} \vec{h}_{t-1} + \vec{b}_{hr} \\
&\vec{Z}_{t} = W_{iz} \vec{x}_{t} + \vec{b}_{iz} + W_{hz} \vec{h}_{t-1} + \vec{b}_{hz} \\
&\vec{M}_{t} = W_{hn} \vec{h}_{t-1} + \vec{b}_{hn}, \quad \vec{N}_{t} = W_{in} \vec{x}_{t} + \vec{b}_{in} + \vec{r}_{t} \circ \vec{M}_{t} \\
&\vec{r}_{t} = \sigma (\vec{R}_{t}), \quad \vec{z}_{t} = \sigma (\vec{Z}_{t}), \quad \vec{n}_{t} = tanh(\vec{N}_{t}) \\
&\vec{h}_{t} = (1 - \vec{z}_{t}) \circ \vec{n}_{t} + \vec{z}_{t} \circ \vec{h}_{t-1}
\end{aligned}
}
\end{equation*}

Given the GRU expressed in the above form, the transposed Jacobian between consecutive hidden states $\frac{\partial \vec{h}_{t}}{\partial \vec{h}_{t-1}}$ can be computed analytically:
\begin{equation}
{
\begin{aligned} \label{eqn:gru_jcb} 
&J_{1} = (\frac{\partial \vec{R}_{t}}{\partial \vec{h}_{t-1}})^{T} = W_{hr}^{T} \\
&\vec{j}_{2} = Diag( (\frac{\partial \vec{r}_{t}}{\partial \vec{R}_{t}})^{T} ) = \vec{r}_{t} \circ (1 - \vec{r}_{t}) \\
&\vec{j}_{3} = Diag( (\frac{\partial \vec{N}_{t}}{\partial \vec{r}_{t}})^{T} ) = \vec{M}_{t} \\
&J_{4} = (\frac{\partial \vec{M}_{t}}{\partial \vec{h}_{t-1}})^{T} = W_{hn}^{T} \\
&\vec{j}_{5} = Diag( (\frac{\partial \vec{N}_{t}}{\partial \vec{M}_{t}})^{T} ) = \vec{r}_{t} \\
&\vec{j}_{6} = Diag( (\frac{\partial \vec{n}_{t}}{\partial \vec{N}_{t}})^{T} ) = 1 - \vec{n}_{t} \circ \vec{n}_{t} \\
&\vec{j}_{7} = Diag( (\frac{\partial \vec{h}_{t}}{\partial \vec{n}_{t}})^{T} ) = 1 - \vec{z}_{t} \\
&J_{8} = (\frac{\partial \vec{Z}_{t}}{\partial \vec{h}_{t-1}})^{T} = W_{hz}^{T} \\
&\vec{j}_{9} = Diag( (\frac{\partial \vec{z}_{t}}{\partial \vec{Z}_{t}})^{T} ) = \vec{z}_{t} \circ (1 - \vec{z}_{t}) \\
&\vec{j}_{10} = Diag( (\frac{\partial \vec{h}_{t}}{\partial \vec{z}_{t}})^{T} ) = \vec{h}_{t-1} - \vec{n}_{t} \\
&J_{11} = (\frac{\partial \vec{h}_{t}}{\partial \vec{h}_{t-1}})_{\text{direct}}^{T} = I \circ \vec{z} \\
&\begin{aligned}
\frac{\partial \vec{h}_{t}}{\partial \vec{h}_{t-1}} = &(J_{1} \circ (\vec{j}_{2} \circ \vec{j}_{3})^T + J_{4} \circ \vec{j}_{5}^T) \circ (\vec{j}_{6} \circ \vec{j}_{7})^{T} + \\
&J_{8} \circ (\vec{j}_{9} \circ \vec{j}_{10})^{T} + J_{11}
\end{aligned}
\end{aligned}
}
\end{equation}
where $Diag(.)$ represents taking the diagonal of a square matrix, and $\circ$ represents the broadcasting element-wise (Hadamard) product. Since cuDNN's GRU implementation \cite{cudnn_rnn_paper} is closed source, we cannot access the values of the gates ($\vec{r}_{t}$, $\vec{z}_{t}$, $\vec{n}_{t}$). Therefore, we have to recompute the gates (but in a more parallelized way) during the forward pass in order to compute $\frac{\partial \vec{h}_{t}}{\partial \vec{h}_{t-1}}$ as shown in Equations~\ref{eqn:gru_jcb}. This engineering challenge results in significant \emph{overhead} during the forward pass in our experiments, however, can potentially be resolved if cuDNN's source code were publicly available and modifiable.
\vspace{-0.2cm}
\section{GRU Training Curve} \label{append:gru_training_curve}

\begin{figure}[H]
    \vspace{-0.0cm}
    \centering
    \includegraphics[width=\linewidth]{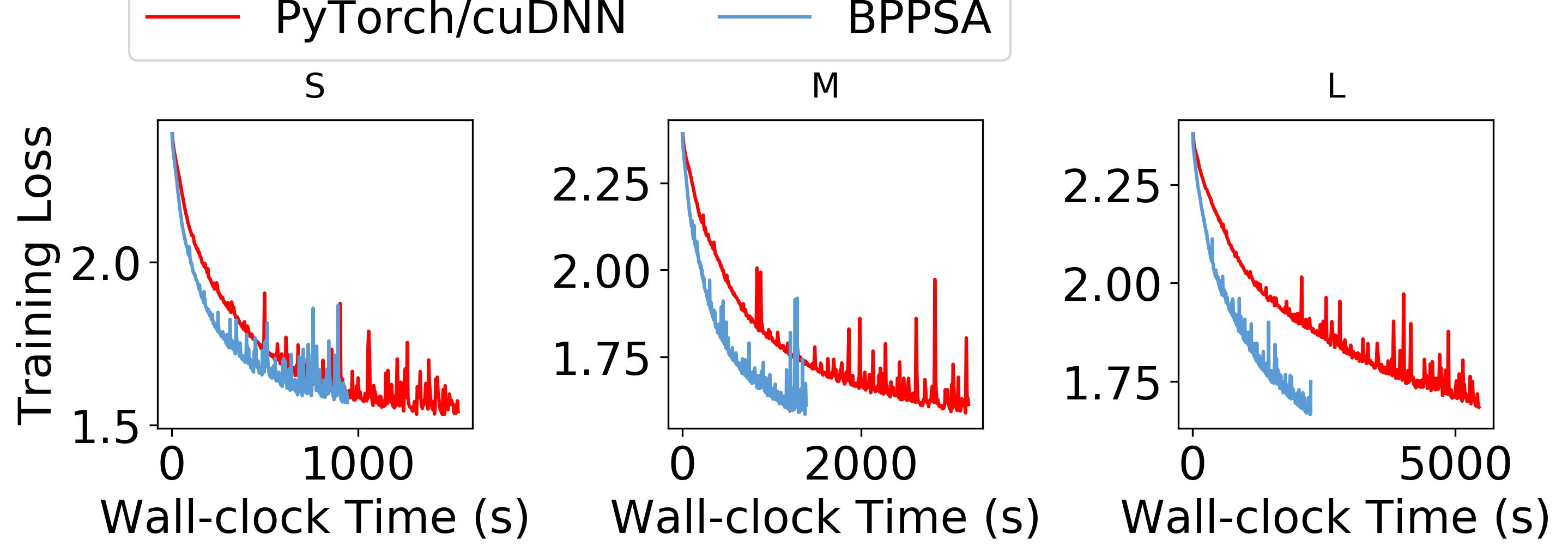}\vspace{-0.3cm}
    \caption{Training loss across wall-clock time when the GRU is trained via BPPSA (blue curve) and the PyTorch Autograd baseline with cuDNN's RNN backend (red curve).}
    \label{fig:gru_training_loss}
    \vspace{-0.5cm}
\end{figure}

Figure~\ref{fig:gru_training_loss} shows the training curves of loss values with respect to wall-clock time when we train the GRU with the (\emph{S}, \emph{M}, \emph{L}) preprocessed datasets for 400 epochs on the RTX 2070 GPU when the mini-batch size $B$ is 16. We observe that the blue curve (BPPSA), if horizontally-scaled, maintains a similar shape as the red curve (PyTorch/cuDNN baseline), which reinforces our observation in Section~\ref{sec:end_to_end_eval} that BPPSA reconstructs the original back-propagation algorithm but achieves a shorter training time.
\vspace{-0.2cm}
\section{Additional Hardware Sensitivity Results} \label{append:v100_results}

To further validate BPPSA on a different GPU architecture, we repeat the experiments in Section~\ref{sec:methodology_end2end} and Section~\ref{sec:methodology_end2end_gru} on an NVIDIA V100 (Volta architecture) \cite{v100} through an AWS p3.2xlarge instance \cite{aws_p3_2xlarge} with the same software stack as our RTX 2080Ti platform (in Table~\ref{tab:spec}). The results are summarized in:
\begin{itemize}
    \item Table~\ref{tab:v100_seq_length_runtimes} and Table~\ref{tab:v100_batch_size_runtimes} for the RNN end-to-end benchmark (Section~\ref{sec:methodology_end2end}). We can derive the backward pass runtime of the \emph{baseline} by subtracting the ``Forward Pass Only" column from the ``Baseline" column for each row ($T$ or $1/B$); while we can also derive the backward pass runtime of \emph{our} method by subtracting the ``Forward Pass Only" column from the ``BPPSA" column for each row ($T$ or $1/B$).
    \item Table~\ref{tab:v100_gru} for the GRU end-to-end benchmark (Section~\ref{sec:methodology_end2end_gru}). We can derive the backward pass runtime of the \emph{baseline} by subtracting the ``\emph{FP}" row from the ``Baseline" row for each column (batch size); while we can also derive the backward pass runtime of \emph{our method} by subtracting the ``\emph{FP} + Overhead" row from the "BPPSA" row for each column (batch size).
\end{itemize}

From the aforementioned results, we can observe trends in both benchmarks similar to the ones in the results from RTX2070 and RTX2080Ti (Section~\ref{sec:end_to_end_eval} and Section~\ref{sec:end_to_end_gru_eval}). 

\begin{table}[h]
\centering
\caption{The wall-clock time (s) for running a single epoch of the RNN end-to-end benchmark (Section~\ref{sec:methodology_end2end}) as the sequence length $T$ increases.} \label{tab:v100_seq_length_runtimes} \vspace{-0.3cm}
\scriptsize
\begin{tabular}{l|l|l|l}
\toprule
\begin{tabular}[c]{@{}l@{}}Sequence\\ Length ($T$)\end{tabular} & \begin{tabular}[c]{@{}l@{}}Forward\\ Pass Only\end{tabular}        & Baseline         & BPPSA           \\ \midrule
10      & $1.57 \pm 0.01$   & $4.49 \pm 0.04$    & $4.24 \pm 0.04$   \\ 
30      & $2.13 \pm 0.01$   & $5.5 \pm 0.06$     & $4.91 \pm 0.06$   \\ 
100     & $3.82 \pm0.02$    & $8.87 \pm 0.05$    & $6.64 \pm 0.08$   \\ 
300     & $8.79 \pm 0.03$   & $18.49 \pm 0.1$    & $11.71 \pm 0.07$  \\ 
1000    & $25.86 \pm 0.12$  & $53.33 \pm 0.39$   & $29.29 \pm 0.18$  \\ 
3000    & $75.33 \pm 0.36$  & $157.11 \pm 0.58$  & $79.48 \pm 0.45$  \\ 
10000   & $250.28 \pm 0.32$ & $510.13 \pm 1.11$  & $265.02 \pm 0.64$ \\ 
30000   & $748.99 \pm 0.71$ & $1557.94 \pm 6.53$ & $764.31 \pm 0.63$ \\ \bottomrule
\end{tabular} \vspace{-0.3cm}
\end{table}

\begin{table}[h]
\centering
\caption{The wall-clock time (s) for running a single epoch of the RNN end-to-end benchmark (Section~\ref{sec:methodology_end2end}) as the fraction of GPU per sample ($1/B$) increases.} \label{tab:v100_batch_size_runtimes} \vspace{-0.3cm}
\scriptsize
\begin{tabular}{l|l|l|l}
\toprule
\begin{tabular}[c]{@{}l@{}}Fraction of GPU\\ per Sample ($1/B$)\end{tabular} & \begin{tabular}[c]{@{}l@{}}Forward\\ Pass Only\end{tabular}           & Baseline          & BPPSA             \\ \midrule
1/256        & $2.13 \pm 0.02$   & $4.02 \pm 0.06$   & $3.72 \pm 0.02$   \\
1/128        & $3.87 \pm 0.02$   & $7.96 \pm 0.11$   & $5.49 \pm 0.01$   \\
1/64         & $7.51 \pm 0.06$   & $14.77 \pm 0.2$   & $9.27 \pm 0.02$   \\
1/32         & $13.62 \pm 0.11$  & $29.51 \pm 0.34$  & $15.0 \pm 0.12$   \\
1/16         & $25.86 \pm 0.12$  & $53.33 \pm 0.39$  & $29.29 \pm 0.18$  \\
1/8          & $51.29 \pm 0.18$  & $102.77 \pm 0.61$ & $56.37 \pm 0.35$  \\
1/4          & $100.55 \pm 0.25$ & $209.67 \pm 0.66$ & $112.23 \pm 0.47$ \\
1/2          & $200.65 \pm 0.24$ & $409.33 \pm 0.67$ & $227.13 \pm 0.74$ \\ \bottomrule
\end{tabular}\vspace{-0.3cm}
\end{table}

\begin{table}[h]
\centering
\caption{The wall-clock time (s) of running one epoch in the GRU end-to-end benchmark (Section~\ref{sec:methodology_end2end_gru}) as the dataset type (\emph{S}, \emph{M}, \emph{L}) and the batch size $B$ varies. ``\emph{FP}" represents running the forward pass only; ``\emph{FP} + Overhead" represents running the forward pass with GRU's Jacobian generation overhead; ``Baseline" represents training normally via the BP baseline; ``BPPSA" represents training via our method.} \label{tab:v100_gru} \vspace{-0.3cm}
\scriptsize
\begin{tabular}{l|l|l|l|l}
\toprule
\multicolumn{2}{l|}{}                                                                           & \begin{tabular}[c]{@{}l@{}}Batch Size\\ $B = 16$\end{tabular} & \begin{tabular}[c]{@{}l@{}}Batch Size\\ $B = 32$\end{tabular} & \begin{tabular}[c]{@{}l@{}}Batch Size\\ $B = 64$\end{tabular} \\ \midrule
\multirow{4}{*}{\emph{S}} & \emph{FP} & $1.66 \pm 0.01$     & $0.91 \pm 0.01$     & $0.52 \pm 0.0$      \\  
                          & Baseline                                                             & $3.71 \pm 0.02$     & $1.9 \pm 0.01$      & $1.0 \pm 0.0$       \\  
                          & \emph{FP} + Overhead & $2.1 \pm 0.01$      & $1.09 \pm 0.01$     & $0.62 \pm 0.0$      \\  
                          & BPPSA                                                                & $2.94 \pm 0.02$     & $1.48 \pm 0.01$     & $0.82 \pm 0.01$     \\ \midrule
\multirow{4}{*}{\emph{M}} & \emph{FP}          & $3.21 \pm 0.01$     & $1.73 \pm 0.01$     & $0.86 \pm 0.01$     \\  
                          & Baseline                                                             & $6.19 \pm 0.04$     & $3.3 \pm 0.02$      & $1.82 \pm 0.02$     \\  
                          & \emph{FP} + Overhead & $3.52 \pm 0.02$     & $1.83 \pm 0.01$     & $1.18 \pm 0.01$     \\  
                          & BPPSA                                                                & $4.2 \pm 0.03$      & $2.22 \pm 0.01$     & $1.2 \pm 0.01$      \\ \midrule
\multirow{4}{*}{\emph{L}} & \emph{FP}          & $5.78 \pm 0.03$     & $3.04 \pm 0.02$     & $1.64 \pm 0.01$     \\  
                          & Baseline                                                             & $11.43 \pm 0.09$    & $6.09 \pm 0.07$     & $3.19 \pm 0.04$     \\  
                          & \emph{FP} + Overhead & $6.06 \pm 0.04$     & $3.23 \pm 0.02$     & $2.13 \pm 0.01$     \\  
                          & BPPSA                                                                & $6.91 \pm 0.05$     & $3.57 \pm 0.04$     & $2.33 \pm 0.01$     \\ \bottomrule
\end{tabular}\vspace{-0.3cm}
\end{table}

%


\end{document}